%% file: main.tex
\documentclass{article}

\PassOptionsToPackage{numbers, compress}{natbib}

\usepackage[preprint]{neurips_2025}
\usepackage{xspace}

\usepackage{color}
\usepackage{colortbl}
\usepackage{xcolor}
\definecolor{citecolor}{HTML}{0071BC}
\definecolor{linkcolor}{HTML}{ED1C24}

\usepackage[utf8]{inputenc} 
\usepackage[T1]{fontenc}    
\usepackage[pagebackref, breaklinks=true, colorlinks,citecolor=citecolor, linkcolor=linkcolor, bookmarks=false]{hyperref}
\usepackage{url}            
\usepackage{booktabs}       
\usepackage{amsfonts}       
\usepackage{nicefrac}       
\usepackage{microtype}      
\usepackage{xcolor}         
\usepackage{fancyvrb}
\usepackage{listings}
\usepackage{multirow}
\usepackage{algorithm}
\usepackage[noEnd=false,indLines=true]{algpseudocodex}
\usepackage{bm}
\usepackage{bbm}
\usepackage{amsmath}
\usepackage{wrapfig}
\usepackage{graphicx}
\usepackage{colortbl}
\usepackage{array}
\usepackage[utf8]{inputenc}
\usepackage{graphicx}
\usepackage{booktabs}
\usepackage{bbm}
\usepackage{wrapfig}
\usepackage{booktabs}
\usepackage{lipsum}
\usepackage{caption}
\usepackage{algorithm}
\usepackage[noEnd=false,indLines=true]{algpseudocodex}
\usepackage{booktabs}
\usepackage{bbm}
\usepackage{wrapfig}
\usepackage{booktabs}
\usepackage{lipsum}
\usepackage{caption}
\usepackage{bm}
\usepackage{listings}
\usepackage{amssymb}

\usepackage[accsupp]{axessibility}  

\usepackage{multirow}
\newlength\savewidth
\usepackage[table]{xcolor}
\newcommand\midline{\noalign{\global\savewidth\arrayrulewidth
		\global\arrayrulewidth 0.5pt}\hline\noalign{\global\arrayrulewidth\savewidth}}
\definecolor{mycolor_blue}{HTML}{FFFFFF}
\definecolor{mycolor_green}{HTML}{EEEEEE}
\definecolor{mycolor_gray}{HTML}{FFFFFF}
\definecolor{myblue}{HTML}{4A90E2}

\definecolor{mycolor_blue}{HTML}{FFFFFF}
\definecolor{mycolor_green}{HTML}{EEEEEE}
\definecolor{mycolor_gray}{HTML}{FFFFFF}
\definecolor{pearDark}{HTML}{2980B9}

\usepackage{tcolorbox}
\usepackage{fbox}
\definecolor{textcolor1}{rgb}{0.25,0.5,0.5}
\definecolor{textcolor2}{rgb}{0.7,0.25,0.25}
\definecolor{linkc}{rgb}{0, 0.44, 0.74}
\definecolor{eqc}{rgb}{1, 0, 0}
\definecolor{myy}{RGB}{126,95,0}
\definecolor{mygray}{gray}{.9}
\definecolor{bblue}{RGB}{30,80,120}
\definecolor{mygray1}{gray}{.7}
\definecolor{ggray}{RGB}{127,127,127}
\definecolor{mygreen}{RGB}{93,174,86}
\definecolor{scolor}{RGB}{111,168,220}
\definecolor{hcolor}{RGB}{111,176,81}
\definecolor{ocolor}{RGB}{224,103,102}
\definecolor{wcolor}{RGB}{246,178,107}
\definecolor{citecolor}{HTML}{229954}
\definecolor{linkcolor}{RGB}{237,4,140}

\def\methodNAME{GRN\xspace}
\def\methodNAMEcompact{GRN}

\title{Generative Refinement Networks for Visual Synthesis}

\vspace{-5pt}
\author{
  \vspace{-25pt}\\
  \textbf{Jian Han,\quad Jinlai Liu,\quad Jiahuan Wang,\quad Bingyue Peng,\quad Zehuan Yuan\thanks{Corresponding author: \href{mailto:yuanzehuan@bytedance.com}{\color{black}{yuanzehuan@bytedance.com}}}}\vspace{5pt} \\
  ByteDance\vspace{3pt} \\
  \texttt{\small \{hanjian.thu123,liujinlai.licio\}@bytedance.com}\\ 
  \texttt{\small \{wangjiahuan.123,bingyue.peng,yuanzehuan\}@bytedance.com}\vspace{4pt}  \\
  Code and models:~\, \url{https://github.com/bytedance/GRN}
  \vspace{-4pt} \\
}

\begin{document}
\maketitle
\input{sec/0_abstract}    
\input{sec/1_intro}

\input{sec/2_related_works}

\input{sec/3_method}

\input{sec/4_experiment}
\input{sec/5_conclusion}

{
\small
\bibliographystyle{cite}
\bibliography{cite}
}
\appendix
\newpage
\input{sec/6_appendix}

\end{document}

%% file: sec/0_abstract.tex
\begin{figure}[ht]
	\centering
	\includegraphics[width=0.86\linewidth]{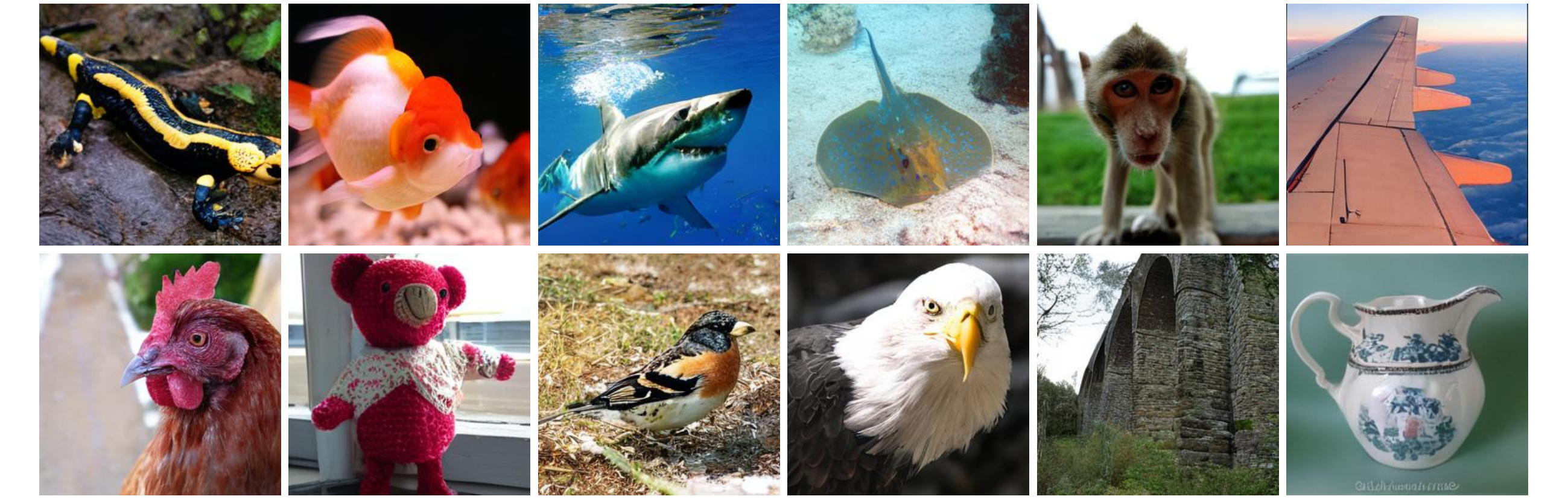}
	\vspace{-0.0cm}
	\caption{\textbf{Qualitative results for the class-to-image generation task.} }
	\label{fig:c2i_examples}
\end{figure}

\begin{abstract}

    While diffusion models dominate the field of visual generation, they remain computationally inefficient, as they allocate uniform computational effort to samples with varying levels of complexity. In contrast, autoregressive (AR) models are inherently complexity-aware, as evidenced by their variable likelihoods, but are often hindered by lossy discrete tokenization and  error accumulation. In this work, we introduce Generative Refinement Networks (\methodNAME), a next-generation visual synthesis paradigm that addresses these issues. At its core, \methodNAME addresses the discrete tokenization bottleneck through a theoretically near-lossless Hierarchical Binary Quantization (HBQ), achieving a reconstruction quality comparable to continuous counterparts. Built upon HBQ's latent space, \methodNAME fundamentally upgrades AR generation with a global refinement mechanism that progressively perfects and corrects artworks — like a human artist painting. Besides, \methodNAME integrates an entropy-guided sampling strategy, enabling complexity-aware, adaptive-step generation without compromising visual quality. On the ImageNet benchmark, \methodNAME establishes new records in image reconstruction (0.56 rFID) and class-conditional image generation (1.81 gFID). We also scale \methodNAME to more challenging text-to-image and text-to-video generation, delivering superior performance on an equivalent scale. We release all models and code to foster further research on \methodNAME.

\end{abstract}

%% file: sec/1_intro.tex
\section{Introduction}
\label{sec:intro}

The field of visual generation has advanced rapidly, driven primarily by scaling diffusion transformers \cite{dit,sora,hunyuanvideo,waver,alive}. By progressively integrating trajectories along a learned velocity field that transports simple noise prior to the empirical data distribution, these models demonstrate strong capabilities in synthesizing high-quality {visual content}.
However, this continuous flow paradigm inherently lacks adaptive-step capacity. Optimized via mean squared error (MSE) without explicit likelihoods, these models are restricted to a fixed number of steps, rigidly allocating identical computational resources to all samples regardless of varying levels of complexity.

Meanwhile, inspired by the success of token-level likelihood estimation in large language models\cite{gpt3.5,gpt4}, autoregressive (AR) models have also garnered extensive research interest in visual synthesis\cite{videogpt,keyuVAR,hanjInfinity,wang2024emu3}. Nevertheless, current AR approaches are bottlenecked by two critical shortcomings. First, they intrinsically suffer from inferior reconstruction quality when utilizing discrete tokens as opposed to continuous representations. Second, their strictly causal prediction mechanism, whether they 
operate  token-by-token or scale-by-scale, 
inevitably causes severe error accumulation over multi-step generation. This exposes a critical lack of error-correction capability, as the model cannot retroactively refine previous mistakes. 
Furthermore, even with parallel prediction in masked AR models \cite{maskgit,bert}, high-confidence tokens become immutable and cannot be revised later. Consequently, such models still inherently lack a holistic refinement mechanism.

These observations motivate a simple yet intuitive refinement-based AR framework with adaptive computation. To this end, we introduce Generative Refinement Networks (\methodNAME), a next-generation visual synthesis paradigm designed to overcome the rigidly fixed computational costs of diffusion models and the inherent shortcomings of standard autoregressive models. Specifically, to address the inferior reconstruction of discrete tokens, we first propose Hierarchical Binary Quantization (HBQ). By ensuring an exponential decay of reconstruction error without increasing latent channels, HBQ empowers discrete image and video tokenizers to achieve near-lossless reconstruction, matching the performance of continuous tokenizers at a higher compression rate. Building upon these robust representations, \methodNAME executes a complexity-aware adaptive-step generation process by employing an entropy-guided sampling mechanism. It dynamically distributes computational loads based on the varying difficulty of visual content, while employing a global refinement mechanism to retroactively mitigate accumulated errors.

Extensive experiments {on diverse visual tasks} validate the superiority of our framework. On the ImageNet 256$\times$256 benchmark \cite{imagenet} for class-conditional image synthesis, \methodNAME sets a new record for both image reconstruction and generation quality. Furthermore, demonstrating exceptional task generalization and scalability, we also successfully scale \methodNAME to high-resolution text-to-image (T2I) and text-to-video (T2V) scenarios. When scaled up, \methodNAME demonstrates the capability to generate photorealistic 1024$\times$1024 images alongside dynamic, high-fidelity 480p videos ranging from 2 to 10 seconds. In summary, our main contributions are as follows:

\begin{enumerate}
	\item We propose \methodNAME, the next-generation visual synthesis framework. It is characterized by a global refinement mechanism and complexity-aware generation, achieving robust and efficient visual generation.
	
	\item We introduce Hierarchical Binary Quantization and contribute a series of discrete image/video tokenizers. For the first time, discrete visual tokenizers are on par with continuous ones with the same latent dimensions.
	
	\item Extensive experiments show that \methodNAME achieves state-of-the-art results on standard C2I benchmarks, with an rFID of 0.56 and a gFID of 1.81. When scaled to more challenging T2I and T2V tasks, it demonstrates superior performance compared to methods at an equivalent scale.
\end{enumerate}

%% file: sec/2_related_works.tex
\section{Related Work}
\label{sec:related_works}

\subsection{Visual Tokenizer}
Visual tokenizers \cite{ldm, vqvae, vqgan} compress visual content for efficient generation. Early vector quantization methods \cite{vqvae,vqgan} map continuous features to a discrete codebook, but suffer from limited scalability, prompting lookup-free approaches \cite{BSQ,fsq}  to enable larger vocabularies. 
Despite this, a performance gap to continuous representations remains. 
Recent works \cite{hanjInfinity,bitdance} aim to close this gap by drastically scaling vocabularies, outperforming continuous VAEs.
However, this gain comes at the cost of slower convergence and larger generative models, motivating more efficient quantization schemes.

\subsection{Autoregressive Models}
Inspired by large language models, \cite{vqgan,llamagen,videopoet,wang2025editinfinity} explore visual generation via next-token prediction. MaskGIT \cite{maskgit} accelerates generation using parallel decoding, where it first generates high-confidence tokens and then iteratively fills in the remainder. VAR \cite{keyuVAR} shifts autoregression to next-scale prediction, improving quality and achieving over $10\times$ faster inference. Nevertheless, AR models remain limited by lossy discrete tokenization and error accumulation, and still lag behind diffusion methods. Although Infinity \cite{hanjInfinity} introduces self-correction by randomly flipping bitwise tokens, its assumption of less than 30\% diffuse errors covers limited patterns.

\subsection{Adaptive-step Generation}
Diffusion models dominate visual generation \cite{FLUX, sdxl, stable-diffusion3,sora, Wan}, but typically require tens of inference steps. Distillation methods \cite{dmd, dmd2} reduce sampling steps substantially, yet still rely on predefined schedules with fixed steps. This ``one-size-fits-all'' strategy wastes computational resources on simple prompts. Recently, AdaDiff \cite{adadiff} employs an external network to determine instance-specific steps and uses a policy gradient method to maximize the reward. The sophisticated pipeline requires an additional network and reward signals.

%% file: sec/3_method.tex
\section{Method}
\label{sec:method}

\begin{figure*}[t]
	\centering
	\includegraphics[width=1.0\linewidth]{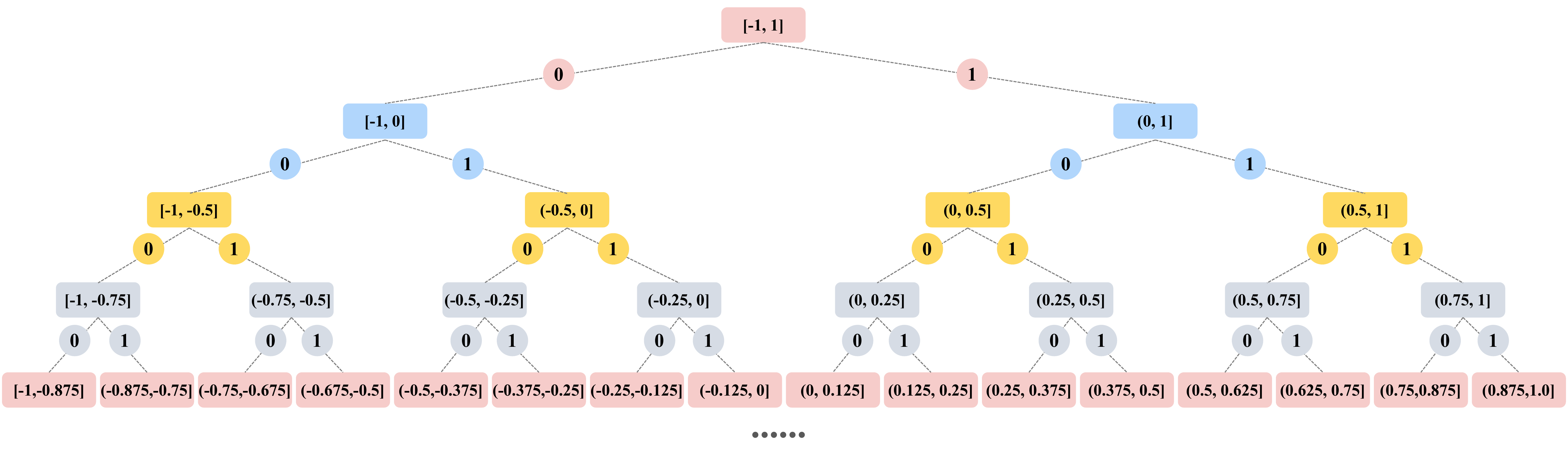}
	\vspace{-0.5cm}
	\caption{\textbf{Hierarchical Binary Quantization.} Each element from the VAE encoded features undergoes several rounds of hierarchical binary quantization. The quantization error decays exponentially with the number of rounds, theoretically enabling lossless quantization to be achieved rapidly.}
	\label{fig:hbq}
\end{figure*}

\subsection{Visual Tokenizer}
The visual tokenizer plays a vital role in learning a compact latent space to compress high-dimensional realistic data. We adopt the 3D causal VAE  design proposed in Wan 2.1 \cite{Wan} so that images and videos can be tokenized in a unified framework.  Specifically, given an image or a video $X \in R^{(1+4T)\times H \times W \times 3}$, the tokenizer encodes its spatio-temporal information into dimensions $[1+T, H/16, W/16]$ while expanding the number of channels to $C$.  Since our goal is lossless discrete compression, and VAE features are continuous signals, we frame feature quantization as a signal transformation problem. Inspired by Harr wavelet \cite{haar1910theorie} in signal processing, we introduce Hierarchical Binary Quantization to transform VAE features into discrete ones. 

\begin{figure*}[t]
	\centering
	\includegraphics[width=1.0\linewidth]{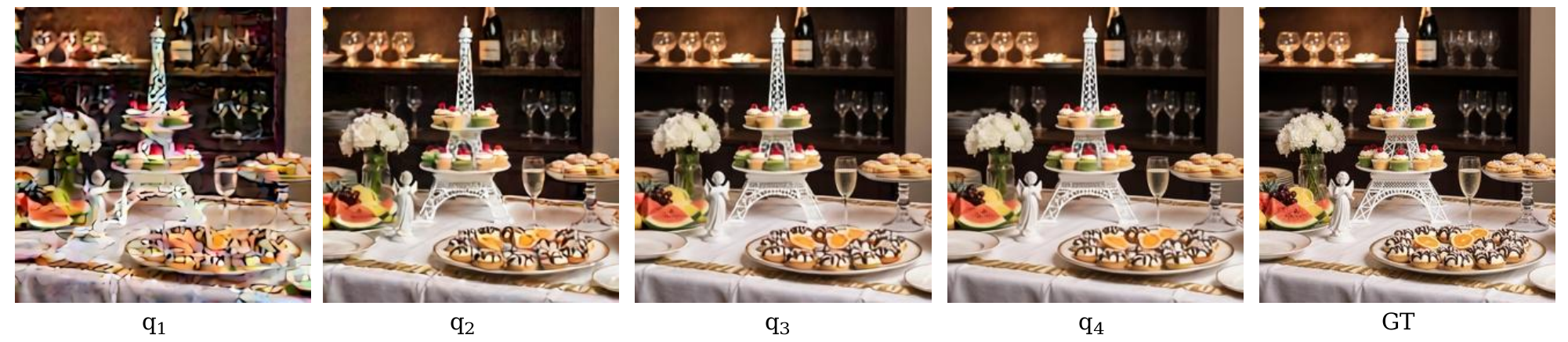}
	\vspace{-0.4cm}
	\caption{\textbf{An example of Hierarchical Binary Quantization (M=4).} For $q_1$, $q_2$, and $q_3$, we truncate the complete sequence and take the truncated parts for reconstruction.}
	\label{fig:hbq_visualize}
\end{figure*}

\textbf{Hierarchical Binary Quantization.} We first append a $\tanh(\cdot)$ non-linear activation function after the VAE encoder to map the feature representation $F$ from an unbounded range to the closed interval $(-1, +1)$. As illustrated in Fig. \ref{fig:hbq}, each element in $F$ undergoes several rounds of binary quantization  based on a binary tree of  buckets with center $c$  as defined in Eq. \ref{eq:base} and Eq. \ref{eq:sign}. 
\begin{equation}
	c_i = \sum^{i-1}_{j=1}  \frac{\delta [q_j]}{2^{j}}
	\label{eq:base}
\end{equation}

\begin{equation}
	q_i=
	\begin{cases} 
		0 & \text{if } F \le c_i, \\
		1 & \text{if } F > c_i.
	\end{cases}
	\label{eq:sign}
\end{equation}
where $\delta(\cdot)$ is a delta function with -1 when $q_i=0$ and 1 otherwise. Then we obtain the quantized binary labels $\{q_1, q_2, ..., q_{M}\}$, where $q_j \in \{0, 1\}^{[1+T, H/16, W/16, C]}$. Here $M$ is the total number of rounds of hierarchical binary quantization.  In this way, we perform quantization from coarse to fine to represent information of different frequencies, and the quantization error $e_j$ for the round $j$ is less than $\frac{1}{2^{j}}$. The upper bound of the quantization error decays exponentially with the number of rounds, theoretically enabling lossless quantization to be achieved rapidly. Fig. \ref{fig:hbq_visualize} shows the images reconstructed from the quantized intermediate results, revealing the coarse-to-fine property. 

\begin{equation}
	\hat{F} = \delta [q_1] \cdot 2^{-1} +  \delta [q_2] \cdot 2^{-2} + ... +  \delta [q_M] \cdot 2^{-M}
	\label{eq:quant_feature}
\end{equation}

Subsequently, the quantized feature $\hat{F}$ can be derived according to Eq. \ref{eq:quant_feature}. The detailed algorithm for HBQ is provided in Appendix \ref{appendix:hbq_alg}. During the training phase of the visual tokenizer, the quantized feature $\hat{F}$ is taken as input to the decoder to reconstruct the raw image or video $X$. Following the common practice for training discrete visual tokenizers, we adopt the Straight-Through Estimator (STE) to backpropagate gradients to the encoder. The training loss is a weighted combination of the reconstruction loss ($\lambda_{recons}$), LPIPS perceptual loss ($\lambda_{LPIPS}$), and the GAN loss ($\lambda_{GAN}$) from a PatchGAN discriminator.

After tokenization, we obtain binary outputs with the size $[1+T, H/16, W/16,C,M]$. However, it is nearly impossible to encode outputs into INT scalars to perform generation by merging $C$ and $M$ dimensions due to its equivalence to introducing a codebook with the large size $2^{CM}$. Inspired by bitwise tokens \cite{hanjInfinity}, we propose two variants of \methodNAME, \emph{i.e.} \methodNAMEcompact$_{ind}$ and \methodNAMEcompact$_{bit}$ to support generation. For \methodNAMEcompact$_{ind}$ we simply encode the $M$ dimension to INT scalars, resulting in $Y_{ind} \in \{0,...,2^M-1\}^{[1+T, H/16, W/16,C]}$. For \methodNAMEcompact$_{bit}$ we concatenate the last two dimensions and predict $Y_{bit} \in \{0,1\}^{[1+T, H/16, W/16,CM]}$. For both variants, we flatten the spatiotemporal dimensions and predict the entire channel dimension in parallel for each token via multi-token prediction \cite{deepseek_v3}.

\subsection{Generative Refinement Network}

Inspired by the intuition of human drawing, we propose an elegantly simple autoregressive refinement framework for visual generation, which commences with a random token map. Let $F_t$ represent the state of the token map in step $t$. The objective is to predict the drawing map $Y_{t+1}$, based on the current state $F_t$. To explicitly formulate this process, we define $F_t$ as a composition of three components: a random map $Y_{rand}$, a drawing map $Y_t$, and a binary selection map $S_t$. The relationship is formally expressed in Eq. \ref{eq:ft}, where $F_t$ is constructed by selecting from $Y_t$ or $Y_{rand}$ based on the values in $S_t$.

\begin{figure*}[t]
	\centering
	\includegraphics[width=1.0\linewidth]{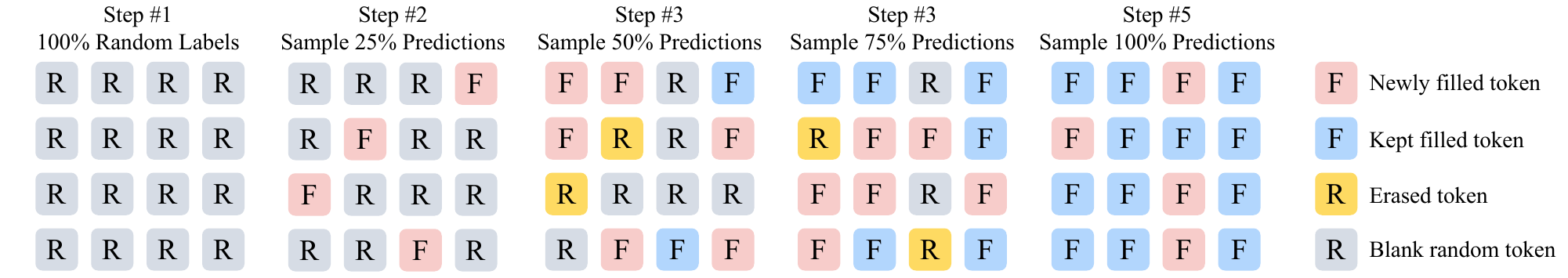}
	\vspace{-0.5cm}
	\caption{\textbf{Generative Refinement Networks.} Starting from a random token map, \methodNAME randomly selects more predictions at each step and refines all input tokens. For example, compared to the second step, the third step filled six new tokens (\textcolor[RGB]{220, 120, 117}{pink}), kept two tokens (\textcolor[RGB]{88, 160, 227}{blue}), erased two tokens (\textcolor[RGB]{240, 180, 40}{yellow}), and left six tokens blank (\textcolor[RGB]{128, 138, 151}{gray}).}
	\label{fig:framework}
\end{figure*}

\begin{equation}
	F_{t} = S_t \cdot Y_t \oplus  \overline{S_t} \cdot Y_{rand} 
	\label{eq:ft}
\end{equation}

Intuitively, $S_t \cdot Y_t$ represents the current drawing while $\overline{S_t} \cdot Y_{rand} $ corresponds to the blank area without any information, mimicking the intermediate step during the human drawing procedure.  $S$ is designed to make the accumulation statistic $l$, namely the proportion of ones in $S$, increase monotonically from 0\% to 100\% during the refinement steps. Therefore, $F$ gradually converges to ideal token maps. In order to obtain $Y_{t+1}$, we employ a transformer $\Phi(\cdot)$ and approximate $p(Y_{t+1})$ as
\begin{equation}
	p(Y_{t+1}) = \Phi ( {F}_{t}, cond)
	\label{eq:framwork}
\end{equation}
to model the next-step drawing by fitting real token maps, where $cond$ denotes the generation condition, such as class embeddings or texts. $S_{t+1}$ is constructed following Eq. \ref{eq:mt_pt} based on $Y_{t+1}$. An alternative approach based on prediction confidence was also investigated for constructing $S_{t+1}$. However, it produced inferior results, as detailed in Appendix \ref{appendix:random_confidence}.

\begin{equation}
	S_t = RandLike(Y_t)<l_t
	\label{eq:mt_pt}
\end{equation}
We introduce a complexity-aware sampling strategy to control $l_{t}$ to not only maintain its monotonicity, but also consider the uncertainty of $Y$. We will discuss the details below. Based on the prediction $Y_{t+1}$ and selection map $S_{t+1}$,  the state $F_{t+1}$ is updated accordingly as mentioned in Eq. \ref{eq:ft}. Thus, a coherent loop of progressive generation and refinement is formed. The autoregressive mechanism allows us not only to improve an increasing number of tokens with high certainty, but also to erase obvious errors with more context included as the drawing proceeds. Ideally, the process converges to the best result as more and more information accumulates. The toy process is demonstrated in Fig. \ref{fig:framework}.

\textbf{Training.}
For each iteration in the training stage, we randomly sample random tokens $Y_{rand}$ from a uniform distribution $\{0, 1, ..., 2^M-1\}$ for \methodNAMEcompact$_{ind}$ and $\{0, 1\}$ for \methodNAMEcompact$_{bit}$. The binary map $S_t$ is also uniformly sampled with varying selection ratios that control how many real tokens are used as input. Therefore, the input $F_t$ to the transformer consists of $N\cdot l_t$ tokens sampled from ground-truth tokens $Y_{{gt}}$ and $N\cdot (1-l_t)$ tokens sampled from random tokens $Y_{{rand}}$. Note that token sampling is randomly conducted along the spatial, temporal, and channel dimensions, with no additional priors (Eq. \ref{eq:mt_pt}). $N$ refers to the total number of tokens, which equals $(1+T)\cdot H/16 \cdot W/16 \cdot (C~or~CM)$. Taking $F_t$ constructed by Eq. \ref{eq:ft} as input with partial information, our goal is to predict ground-truth tokens similar to x-prediction in diffusion setting using the simple Cross-Entropy loss as illustrated in Eq. \ref{eq:ce_loss}. Here $y_i$ denotes the ground truth token.

\begin{equation}
	\mathcal{L} = - \mathbf{E} [\frac{1}{N} \sum^{N}_{i=0} \log p(y_i \mid F_t, cond)]
	\label{eq:ce_loss}
\end{equation}

The detailed training and inference process for \methodNAME is provided in Appendix \ref{appendix:grn_alg}. Additional comparisons between \methodNAME and other autoregressive models are provided in Appendix \ref{sec:appendix_ar_diff}.

\textbf{Complexity-Aware Sampling.}
We propose an entropy-guided scheduling function to determine $l_t$, where $t$ refers to the index of the refinement step. In particular, we calculate the average entropy $H(Y_t)$ for step $t$ during generation as

\begin{equation}
	H(Y_t)= \frac{1}{N} \cdot \frac{1}{\log_2 K} \cdot \sum^{N}_{i=0} \sum^{K}_{j=0} -p(y_{(i,j)}\mid F_{t-1}, cond) \cdot \log_2 p(y_{(i,j)} \mid F_{t-1}, cond).
	\label{eq:entropy}
\end{equation}

In Eq. \ref{eq:entropy}, we denote $i$ as the token index and $j\in \{1,2,...,K\}$  as the category index, where $K$ is the total number of categories. Note that $K=2^M$ for \methodNAMEcompact$_{ind}$ and $K=2$ for \methodNAMEcompact$_{bit}$. Generation complexity is measured by the entropy $H(Y_t)$, bounded between 0 and 1. Given that a smaller $H(Y_t)$ denotes greater predictive confidence, we allocate fewer refinement steps alongside a steeper increase in $l_t$, thereby retaining more information from $Y_t$. Conversely, when high entropy indicates substantial complexity, we apply more refinement steps and a moderate progression of $l_t$. Specifically, we formulate $l_t$ as

\begin{equation}
	l_t = l \left( Y,t \right) = \frac{t}{\alpha} \mathbbm{1}_{t \le t_0} + \left( \frac{t_0}{\alpha} + \frac{\alpha-t_0}{\alpha} \cdot \frac{(t-t_0)}{ k\cdot H(Y_{(t_0+1)})+b }\right)  \mathbbm{1}_{t > t_0}.
	\label{eq:pt_func}
\end{equation}

Here, $H(Y_{(t_0+1)})$ represents the average entropy calculated from a specific step. We set a warm-up period with $t_0=5$ and $\alpha=50$, as we observed that the entropy values are unstable during the initial steps. The hyperparameters $k$ control the dynamic range of adaptive steps, and $b$ is the bias.  We also clip the value of $ k\cdot H(Y_{(t_0+1)}) +b$ to ensure the total number of inference steps remains within the range of $[T_{min}, T_{max}]$.

%% file: sec/4_experiment.tex
\begin{table*}[t]
	\centering
	\caption{\textbf{Reconstruction performance comparison of image tokenizers on ImageNet (256$\times$256).}}
	\label{tab:imagenet_recons}
	\resizebox{0.95\linewidth}{!}{ 
		\begin{tabular}{l|ccccc|cccc}
			\toprule
			\multirow{2}{*}{Method} & {Tokenizer}  & Spatial  & Latent & Channel  & Compress & \multirow{2}{*}{rFID$\downarrow$} & \multirow{2}{*}{LPIPS$\downarrow$} & \multirow{2}{*}{SSIM$\uparrow$}  & \multirow{2}{*}{PSNR$\uparrow$} \\ 
			&  Type & Ratio & Channel  & Bits & Ratio  &  &  &  \\
			\midrule
			
			SD-VAE \cite{ldm} & Continuous & 16 & 16 & 16 & 24 & 0.87  & - & 0.68 & 24.08 \\
			RAE \cite{RAE} & Continuous & 16 & 768 & 16 & 0.5 & 0.62   & 0.25 & 0.44 & 19.20 \\
			\midline
			VAR$^\dagger$ \cite{keyuVAR} & Discrete & 16 & N/A & N/A & 193 & 0.85  & 0.15 & 0.64 & 22.47 \\
			LlamaGen \cite{llamagen} & Discrete & 16 & 1 & 14 & 439 & 2.19  &- & 0.68 & 20.79 \\
			Open-MAGVIT2 \cite{open_magvit2}  & Discrete & 16 & 1 & 18 & 341 & 1.17  & - &  - & 22.64 \\
			\cellcolor{mycolor_green}{\textbf{HBQ (M=4)}} & \cellcolor{mycolor_green}{Discrete} & \cellcolor{mycolor_green}{16} & \cellcolor{mycolor_green}{16} & \cellcolor{mycolor_green}{{4}} & \cellcolor{mycolor_green}{96} & \cellcolor{mycolor_green}{\textbf{0.56}} & \cellcolor{mycolor_green}{\textbf{0.13}} & \cellcolor{mycolor_green}{\textbf{0.71}} & \cellcolor{mycolor_green}{23.01} \\			
			\bottomrule
		\end{tabular}
	}
\end{table*}

\begin{table*}[t]
	\centering
	\caption{\textbf{Reconstruction performance of HBQ tokenizers.} To evaluate reconstruction quality, we curated a challenging validation set of 160 high-motion videos. For continuous tokenizers, we consider each latent channel to hold 16 bits of information. An HBQ tokenizer with $M$ rounds introduces $M$ bits within each latent channel. We train HBQ*(M=4) for more iterations and apply an optimized GAN loss weight. Note that \emph{Compress Ratio = (Spatial Stride)$^2$ * Temporal Stride * 3 / Latent Channel * 8 / Channel Bits}.}
	\label{tab:video_recons}
	\resizebox{0.95\linewidth}{!}{ 
		\begin{tabular}{l|cccccc|rccc}
			\toprule
			\multirow{2}{*}{Method} & {Tokenizer}  & Latent & Spatial & Temporal & Channel  & Compress & \multirow{2}{*}{rFVD$\downarrow$} & \multirow{2}{*}{LPIPS$\downarrow$} & \multirow{2}{*}{SSIM$\uparrow$}  & \multirow{2}{*}{PSNR$\uparrow$} \\ 
			&  Type & Channel & Stride & Stride  & Bits & Ratio  &  &  &  \\
			\midrule
			Wan 2.1 & Continuous & 16 & 8 & 4 & 16 & 24 & 19.5  & 0.058 & 0.929 & 34.10 \\
			Wan 2.2 & Continuous & 48 & 16 & 4 & 16 & 32 & 22.6  & 0.052 & 0.932 & 34.54 \\
			\midline
            HBQ (w/o quant)& Continuous & 16 & 16 & 4 & 16 & 96 & 144.6  & 0.141 & 0.879 & 31.14 \\
			HBQ (M=4)& Discrete & 16 & 16 & 4 & 4 & 384 & 163.6  & 0.148 & 0.872 & 30.40 \\
			HBQ (M=6)& Discrete & 16 & 16 & 4 & 6 & 256 & 148.8  & 0.142 & 0.878 & 30.98 \\
			HBQ (M=8)& Discrete & 16 & 16 & 4 & 8 & 192 & 144.9  & 0.141 & 0.879 & 31.10 \\
			\midline
            HBQ (w/o quant) & Continuous & 64 & 16 & 4 & 16 & 24 & 43.2  & 0.078 & {0.935} & {34.79} \\
			HBQ (M=4)& Discrete & 64 & 16 & 4 & 4 & 96 & 50.6  & 0.084 & 0.930 & 33.97 \\
            \rowcolor{mycolor_green}
			HBQ* (M=4)& Discrete & 64 & 16 & 4 & 4 & 96 & 26.3  & 0.068 & 0.938 & 34.73 \\
			\bottomrule
		\end{tabular}
	}
\end{table*}

\begin{figure}[b]
	\centering
	\includegraphics[width=1.0\linewidth]{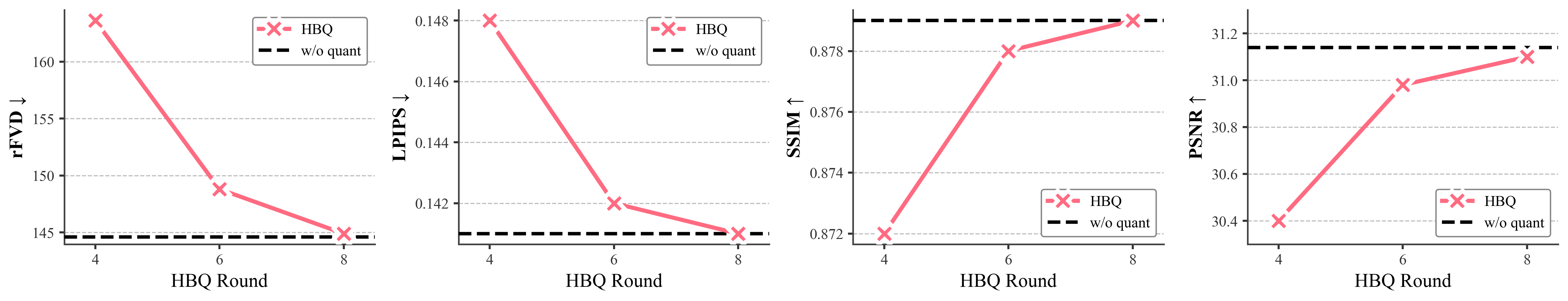}
	\caption{\textbf{Effect of HBQ rounds.} 8-round HBQ matches the continuous baseline.}
	\label{fig:ablation_hb_round}
\end{figure}

\section{Experiments}
\label{sec:experiment}

\subsection{Visual Tokenizer}
\textbf{Implementation.} We introduce two visual tokenizers: an image-only tokenizer tailored for class-conditional image generation, and a joint image-video tokenizer designed for text-to-image and text-to-video generation tasks. Both tokenizers adopt the 3D causal encoder and decoder architecture from Wan 2.1 \cite{Wan} and are trained from scratch. Specifically, the image-only tokenizer is trained on the OpenImages dataset \cite{openimages}, while the joint tokenizer is trained on a combination of publicly available image and video datasets. During training, the overall objective comprises reconstruction, perceptual, and adversarial (GAN) losses. The respective loss weights are set to 1.0, 1.0, and 0.3 for the image-only tokenizer, and 1.0, 0.2, and 0.005 for the joint image-video tokenizer.

\textbf{Results. }As shown in Tab. \ref{tab:imagenet_recons}, our tokenizer achieves state-of-the-art reconstruction performance on the 256$\times$256 ImageNet benchmark. Utilizing four HBQ rounds, it achieves a remarkable rFID of 0.56. This result significantly surpasses not only the continuous SD-VAE (0.87 rFID) while operating at a 4$\times$ higher compression rate, but also substantially outperforms other leading methods, including RAE (0.62), VAR (0.85), LlamaGen (2.19), and Open-MAGVIT-v2 (1.17). These results underscore our method's superior ability to achieve high fidelity reconstruction under stringent compression.

In Tab. \ref{tab:video_recons}, we present a series of joint image-video tokenizers, exploring the impact of varying HBQ rounds and latent channel dimensions. We first analyze the effect of HBQ rounds and observe a clear trend: reconstruction metrics such as rFVD and PSNR consistently improve as the number of HBQ rounds increases. While four to six rounds already yield strong performance, an eight-round configuration achieves reconstruction quality nearly identical to that of the continuous baseline as depicted in Tab. \ref{tab:video_recons} and Fig. \ref{fig:ablation_hb_round}. This demonstrates that our HBQ tokenizer can match its continuous counterpart's fidelity while operating at a higher compression rate. Crucially, this is achieved without increasing the latent channels. While other methods (e.g., Infinity \cite{hanjInfinity}, BitDance \cite{bitdance}) can also bridge the gap to continuous models, they typically rely on expanding the latent dimension. As recent studies \cite{stable-diffusion3, hanjInfinity, dcae1p5} suggest, such an approach often slows convergence and necessitates larger models. In contrast, our primary results are achieved without increasing the latent dimension.

We also experimented with expanding the latent channels from 16 to 64. This single change boosts the PSNR from 30.40 to an impressive 33.97. Notably, this performance is comparable to the state-of-the-art Wan 2.1 tokenizer, but is achieved at a 4$\times$ higher compression rate. By carefully tuning the GAN loss weight and training for more iterations, the HBQ tokenizer outperforms Wan 2.1 and Wan 2.2 in terms of SSIM and PSNR. More details on tuning $\lambda_{GAN}$ are provided in Appendix \ref{appendix:tune_gan}.

\subsection{Class-to-Image Results}
\textbf{Implementation.} Following JiT \cite{jitpaper}, we incorporate SwiGLU, RMSNorm, RoPE, qk-norm, and in-context class conditioning to the original transformer. We train \methodNAMEcompact$_{ind}$ with four different model sizes: 130M, 458M, 952M, and 2B, denoted as \methodNAME-B, \methodNAME-L, \methodNAME-H, \methodNAME-G, respectively. We train them for 600 epochs on the ImageNet \cite{imagenet} dataset. The learning rate is set to 2e-4 and is constant during training. We randomly discard 10\% conditions for Classifier-Free Guidance. During the inference stage, we grid search the best decoding hyperparameters. Additional implementation details are provided in Appendix 4.

\newcommand\headspace{\hspace{.2em}}
\newcommand\shrink[1]{{\fontsize{6pt}{7.2pt}\selectfont{#1}}}

\begin{table}[t]
	\caption{
		\textbf{Reference results on ImageNet 256$\times$256.} 
		FID \cite{fid} and IS \cite{inception_score} of 50K samples are evaluated. Tokenizer ``D'' refers to a discrete tokenizer, while ``C'' refers to a continuous tokenizer.
		\label{tab:in256-sys}
	}
	\begin{tabular}{l l c c c c|c c}
		\midline
		\textbf{Type} & \textbf{Model} & \textbf{Tokenizer} & \textbf{Loss} & \textbf{Param} & \textbf{Gflops} & {\textbf{FID}$\downarrow$} & {\textbf{IS}$\uparrow$} \\
		\midline
		\headspace Diffusion & \headspace DiT-L/2 \cite{dit} & C & MSE & 458M & - & 5.02 & 167.2  \\
		\headspace Diffusion & \headspace DiT-XL/2 \cite{dit} & C & MSE & 675M & 119 & 2.27 & 278.2 \\
		\headspace Flow & \headspace SiT-XL/2 \cite{sit} & C & MSE & 675M  & 119 & 2.06 & 277.5 \\
		\headspace Flow & \headspace REPA \cite{repa}, SiT-XL/2  & C & MSE & 675M & 119 & 1.42 & 305.7 \\
		\headspace Flow & \headspace RAE \cite{RAE}, DiT$^{\text{DH}}$-XL/2 & C & MSE & \hspace{0pt} 839M & 146 & {1.13} & 262.6 \\
		\headspace Flow & \headspace {JiT-B/16} \cite{jitpaper} & C & MSE & 131M  & 25 & 3.66 & 275.1 \\
		\headspace Flow & \headspace {JiT-L/16} \cite{jitpaper} & C & MSE & 459M  & 88 & 2.36 & 298.5 \\
		\headspace Flow & \headspace {JiT-H/16} \cite{jitpaper} & C & MSE & 953M  & 182 & 1.86 & 303.4 \\
		\headspace Flow & \headspace {JiT-G/16} \cite{jitpaper} & C & MSE & 2B & 383 & 1.82 & 292.6 \\
		\midline
		\headspace Hybrid & \headspace {MAR} \cite{MAR} & C & MSE & 943M & - & 1.55 & 303.7 \\
		\headspace Hybrid & \headspace {BitDance-H-1x} \cite{bitdance} & D & MSE & 1B & - & 1.24 & 304.4 \\
		\midline
		\headspace AR & \headspace LlamaGen-L \cite{llamagen} & D & CE &  343M  & - & 3.07 & 256.1 \\
		\headspace AR & \headspace LlamaGen-XL \cite{llamagen} & D & CE &  775M  & - & 2.62 & 244.1 \\
		\headspace AR & \headspace LlamaGen-XXL \cite{llamagen} & D& CE &  1.4B  & - & 2.34 & 253.9 \\
		\headspace AR & \headspace MaskGIT \cite{maskgit} & D& CE & 227M  & - & 6.18 & 182.1\\
		\headspace AR & \headspace VAR-d20 \cite{keyuVAR} & D& CE & 600M  & - & 2.57 & 302.6 \\
		\headspace AR & \headspace VAR-d24 \cite{keyuVAR} & D& CE &  1B  & - & 2.09 & 312.9\\
		\headspace AR & \headspace VAR-d30 \cite{keyuVAR} & D& CE & 2B  & - & 1.92 & 323.1 \\
		\headspace AR & \headspace RandAR-XXL \cite{pang2025randar} & D& CE  &  1.4B  & - & 2.15 & 322.0 \\
		\midline
		\headspace AR & \headspace \textbf{\methodNAMEcompact-B} & D& CE & 130M & 25 & 3.56 & 280.3 \\
		\headspace AR & \headspace \textbf{\methodNAMEcompact-L} & D& CE & 458M & 88 & 2.64 & 314.8 \\
		\headspace AR & \headspace \textbf{\methodNAMEcompact-H} & D& CE & 952M & 182 & 2.06 & 316.1 \\
		\headspace AR & \headspace \textbf{\methodNAMEcompact-G} & D& CE & 2B & 383 & 1.81 & 299.0 \\
		\midline
	\end{tabular}
\end{table}

\textbf{Results. }As shown in Tab. \ref{tab:in256-sys}, \methodNAME is benchmarked against state-of-the-art diffusion, hybrid, and autoregressive models on the ImageNet 256$\times$256 class-conditional generation task. With nearly half the parameters of MaskGIT \cite{maskgit}, our \methodNAME-B model achieves a superior FID of 3.56 (vs. 6.18), demonstrating remarkable efficiency. Our largest variant, \methodNAME-G, achieves a state-of-the-art FID of 1.81, rivaling top diffusion and hybrid models. Notably, \methodNAME-G surpasses foundational models like DiT \cite{dit} and SiT \cite{sit} in both FID and Inception Score. This is significant as they form the backbone of many current industrial T2I and T2V models. Furthermore, \methodNAME-G outperforms the autoregressive model LlamaGen \cite{llamagen} and VAR \cite{keyuVAR}. We attribute this advantage to our proposed global refinement generation, which effectively mitigates error propagation. The high-quality visual samples generated by \methodNAME-G, shown in Fig. \ref{fig:c2i_examples}, also confirm its capabilities. Please refer to Fig. 4 in the Appendix for additional uncurated qualitative results. These strong results establish \methodNAME as a powerful and scalable baseline for high-fidelity visual generation, motivating its application to more complex text-to-image and text-to-video tasks.

\subsection{Text-to-Image  Results}
\textbf{Implementation.} We train \methodNAMEcompact$_{bit}$ for the text-to-image task with 2B parameters from scratch. In contrast to C2I models, the T2I model leverages in-context self-attention instead of adaln-zero to inject conditions. The model was pre-trained on large-scale public datasets and subsequently fine-tuned on a small, high-quality proprietary dataset. We first train \methodNAME on the pre-training dataset at a resolution of 256$\times$256 for 150K iterations using a batch size of around 15,400 and a learning rate of 2e-4. Then we fine-tune \methodNAME at 1024 resolution with a smaller, high-quality dataset. In this stage, we train \methodNAME for 60K iterations using a batch size of 2048 and a learning rate of 2e-5. Additional implementation details are provided in Appendix 4. Furthermore, we observe that GRN significantly outperforms Infinity in generating fine details such as small faces.

\textbf{Results. }As shown in Tab. \ref{tab:geneval_only}, our model, augmented with a rewriter, achieves an overall score of 0.76 on the GenEval benchmark \cite{ghosh2024geneval}. While our model is outperformed by larger-scale methods such as Z-Image-Turbo \cite{cai2025z}, HiDream \cite{cai2025hidream}, Qwen-Image \cite{qwenimage2025report}, and BitDance \cite{ai2026bitdance}, it is crucial to note the significant disparity in model size: these models utilize 6B to 20B parameters, whereas \methodNAME{} is a far more compact 2B model. When compared at an equivalent scale, \methodNAME{} demonstrates superior performance, significantly outperforming models of a similar 2B size like SD3 Medium \cite{stable-diffusion3} (0.62) and Infinity \cite{hanjInfinity} (0.71). The qualitative results in Fig. 5 in the Appendix showcase \methodNAME's strong capability to generate high-fidelity and diverse images that accurately follow user prompts.

\begin{table}[t]
	\caption{\textbf{Evaluation of Text-to-Image generation on GenEval~\cite{ghosh2024geneval}.}}
	\label{tab:geneval_only}
	\centering
	\small
	\setlength{\tabcolsep}{5pt}
	\renewcommand\arraystretch{1.0}
	\resizebox{1.0\linewidth}{!}{
		\begin{tabular}{lccccccccc}
			\toprule
			\textbf{Model} & \textbf{\#Param} &\textbf{\#Data} &\textbf{Single Obj.} & \textbf{Two Obj.} & \textbf{Count} & \textbf{Colors} & \textbf{Pos.} & \textbf{Color Attri.} & \textbf{Overall$\uparrow$} \\
			\midrule
			\multicolumn{7}{l}{\textit{Proprietary Models}} \\
			~ GPT Image 1~\cite{openai2025gpt4oimage}   & N/A  & N/A & 0.99  & 0.92 & 0.85 & 0.92 & 0.75 & 0.61 & 0.84 \\
			~ Seedream 3.0~\cite{gao2025seedream}   & N/A  & N/A & 0.99      & 0.96 & 0.91 & 0.93 & 0.47 & 0.80 & 0.84 \\
			\midrule
			\multicolumn{7}{l}{\textit{Diffusion Models}} \\
			~ PixArt-$\alpha$~\cite{chen2023pixart}  & 0.6B  & N/A & 0.98     & 0.50 & 0.44 & 0.80 & 0.08 & 0.07 & 0.48 \\
			~ SD3 Medium~\cite{stable-diffusion3}        & 2B  & N/A & 0.98      & 0.74 & 0.63 & 0.67 & 0.34 & 0.36 & 0.62 \\
			~ JanusFlow~\cite{janusflow}      & 1.3B  & N/A & 0.97      & 0.59 & 0.45 & 0.83 & 0.53 & 0.42 & 0.63 \\
			~ FLUX.1-Dev~\cite{FLUX}            & 12B  & N/A & 0.98          & 0.81 & 0.74 & 0.79 & 0.22 & 0.45 & 0.66 \\
			~ SD3.5-Large~\cite{stable-diffusion3}        & 8B  & N/A & 0.98     & 0.89 & 0.73 & 0.83 & 0.34 & 0.47 & 0.71 \\
			~ Lumina-Image-2.0~\cite{qin2025lumina}   & 2.6B  & 111M & -    & 0.87 & 0.67 & -    & -    & 0.62 & 0.73 \\
			~ Show-o2~\cite{xie2025show}             & N/A  & N/A & 1.00      & 0.87 & 0.58 & 0.92 & 0.52 & 0.62 & 0.76 \\
			~ Z-Image-Turbo~\cite{cai2025z} & 6B  & N/A & 1.00      & 0.95 & 0.77 & 0.89 & 0.65 & 0.68 & 0.82 \\
			~ HiDream-I1-Full~\cite{cai2025hidream}   & 17B  & N/A & 1.00    & 0.98 & 0.79 & 0.91 & 0.60 & 0.72 & 0.83 \\
			~ Z-Image~\cite{cai2025z}          & 6B   & N/A & 1.00    & 0.94 & 0.78 & 0.93 & 0.62 & 0.77 & 0.84 \\
			~ Qwen-Image~\cite{qwenimage2025report}          & 20B  & N/A & 0.99      & 0.92 & 0.89 & 0.88 & 0.76 & 0.77 & 0.87 \\
			~ BAGEL~\cite{deng2025emerging}             & 14B  & N/A & 0.98   & 0.95 & 0.84 & 0.95 & 0.78 & 0.77 & 0.88 \\
			\midrule
			\multicolumn{7}{l}{\textit{Autoregressive Models}} \\
			~ Emu3-Gen~\cite{wang2024emu3}            & 8B  & N/A & 0.98    & 0.71 & 0.34 & 0.81 & 0.17 & 0.21 & 0.54 \\
			~ Infinity$^{\dagger}$~\cite{hanjInfinity}       & 2B  & 160M & -   & 0.85 & -    & -    & 0.49 & 0.57 & 0.73 \\
			~ Janus-Pro~\cite{chen2025janus}        & N/A  & N/A & {0.99}      & 0.89 & 0.59 & {0.90} & {0.79} & {0.66} & 0.80 \\
			~ Tar~\cite{han2025vision}         & N/A  & N/A & 0.98 & {0.92} & {0.83} & 0.85 & {0.80} & 0.65 & {0.84} \\
			~ NextStep-1~\cite{team2025nextstep}   & 14B  & N/A & -       & -    & -    & -    & -    & -    & 0.73 \\
			~ {BitDance~\cite{ai2026bitdance}}                 & 14B & N/A &  {1.00}  & {0.96} & {0.71} & {0.95} & 0.72 & {0.83} & {0.86} \\
			\rowcolor{mycolor_green}
			~ \textbf{\methodNAME$^{\dagger}$}                 & 2B  & 80M & {0.99}   & {0.90} & {0.72} & {0.84} & 0.52 & {0.60} & {0.76} \\
			\bottomrule
		\end{tabular}
	}
\end{table}

\begin{table*}[t]
	\centering
	\caption{\textbf{VBench evaluation results. $\dagger$ denotes results obtained with prompt rewriting.}}
	\label{tab:vBenchTable}
	\resizebox{1.0\linewidth}{!}{ 
		\begin{tabular}{lcccccccc}
			\toprule
			\multirow{2}{*}{Models} & \multirow{2}{*}{\# Params}  & Human  & \multirow{2}{*}{Scene} & Multiple  & Appear.  & Quality & Semantic & \multirow{2}{*}{\textbf{Overall}}  \\ 
			&  & Action &  & Objects & Style  & Score & Score &  \\
			\midrule
			\multicolumn{9}{l}{Diffusion / Flow Models} \\
			\midrule
			AnimateDiff-V2 \cite{animatediff} & 1.5B & 92.60 & 50.19 & 36.88 & 22.42 & 82.90 & 69.75 & 80.27 \\
			VideoCrafter-2.0 \cite{videocrafter} & 1.5B & 95.00 & {55.29} & 40.66 & {25.13} & 82.20 & 73.42 & 80.44 \\
			OpenSora V1.2 \cite{opensora} & 1.1B & 85.80 & 42.47 & 58.41 & 23.89 & 80.71 & 73.30 & 79.23 \\
			Show-1 \cite{show-1} & 6B & 95.60 & 47.03 & 45.47 & 23.06 & 80.42 & 72.98 & 78.93 \\
			URSA \cite{ursa}  & 1.7B & - & 52.30 & 70.60 & - & 83.40 & 78.50 & 82.40 \\
			CogVideoX-5B \cite{cogvideox} & 5B & {99.40} & 53.20 & 62.11 & {24.91} & 82.75 & 77.04 & 81.61 \\
			HunyuanVideo \cite{hunyuanvideo} & 13B & 94.40 & {53.88} & 68.55 & 19.80 & 85.09 & 75.82 & 83.24 \\
			Wan 2.1 \cite{Wan} & 14B & {98.80} & 53.67 & {81.44} & 21.13 & {85.64} & {80.95} & {84.70} \\
			\midrule
			\multicolumn{9}{l}{AutoRegressive Models} \\
			\midrule
			Nova$^{\dagger}$ \cite{nova} & 0.6B & 95.20 & 54.06 & 77.52 & 20.92 & 80.39 & 79.05 & 80.12 \\
			Emu3 \cite{wang2024emu3} & 8B & 77.71 & 37.11 & 44.64 & 20.92 & 84.09 & 68.43 & 80.96 \\
			Lumos-1 \cite{yuanlumos}  & {3.6B} & - & - & - & - & 79.50 & 73.50 & 78.30 \\
			{{InfinityStar}$^{\dagger}$} \cite{infinitystar} & {8B} & {96.43} & {52.08} & 
			{{78.66}} & {21.81} & {{84.73}} & {{79.78}} & {{83.74}} \\
			\cellcolor{mycolor_green}{\textbf{\methodNAME}$^{\dagger}$} & \cellcolor{mycolor_green}{2B} & \cellcolor{mycolor_green}{93.75} & \cellcolor{mycolor_green}{50.44} & \cellcolor{mycolor_green}{{70.83}} & \cellcolor{mycolor_green}{21.30} & \cellcolor{mycolor_green}{{84.41}} & \cellcolor{mycolor_green}{{77.35}} & \cellcolor{mycolor_green}{{82.99}} \\
			
			\bottomrule
		\end{tabular}
	}
\end{table*}

\begin{figure}[h!]
	\centering
	\includegraphics[width=1.0\linewidth]{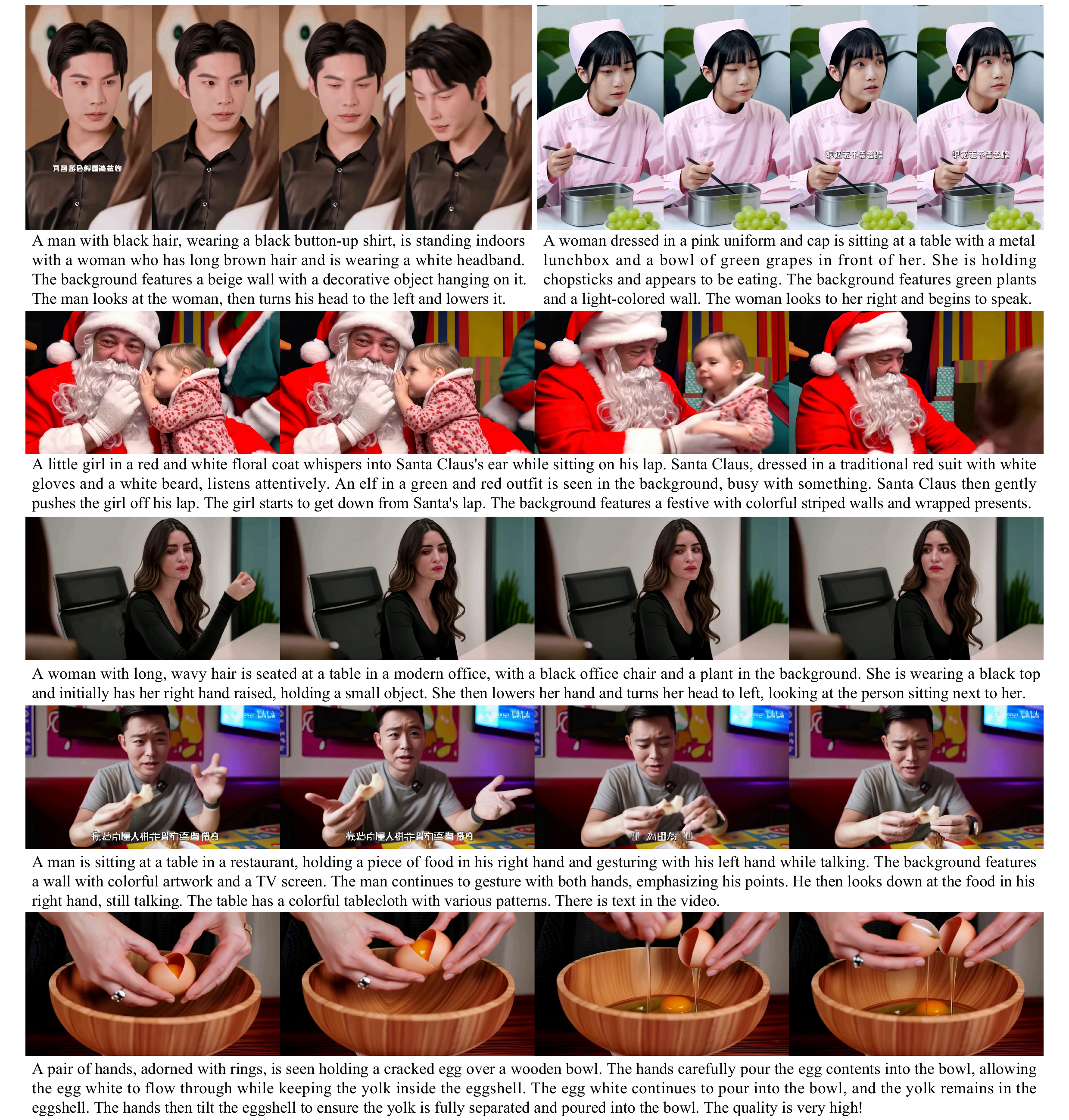}
	\caption{\textbf{Qualitative results of \methodNAME (2B) on the text-to-video task.}}
	\label{fig:t2v_examples}
\end{figure}

\subsection{Text-to-Video Results}
\textbf{Implementation.} Beyond the class-to-image and text-to-image generation tasks, we extend \methodNAME to the most challenging text-to-video synthesis task. The T2V variant of \methodNAME shares the same architecture as its T2I counterpart but is trained exclusively on video data. For this purpose, we curated a training dataset of approximately 40 million video clips, each with a resolution of at least 256$\times$ 256 and a duration of 2 to 10 seconds. The training process consists of two stages. First, we train \methodNAME at a 192p resolution for 150K iterations with a batch size of 4096 and a learning rate of 2e-4. Subsequently, we switch to a 480p resolution for fine-tuning, training for an additional 9K iterations with a reduced batch size of 1350 and a learning rate of 2e-5. Additional implementation details are provided in Appendix 4.

\textbf{Results. }As demonstrated in Tab. \ref{tab:vBenchTable}, \methodNAME exhibits superior performance in generating videos from textual prompts. When benchmarked against contemporary diffusion and flow-based models—including AnimateDiff-V2 \cite{animatediff}, VideoCrafter-2.0 \cite{videocrafter}, OpenSora V1.2 \cite{opensora}, Show-1 \cite{show-1}, and CogVideoX-5B \cite{cogvideox}—\methodNAME  achieves significantly higher scores across quality, semantic, and overall scores. Notably, despite having only 2B parameters, \methodNAME surpasses the much larger CogVideoX-5B \cite{cogvideox} model, highlighting its exceptional parameter efficiency. Furthermore, our approach outperforms URSA \cite{ursa}, a discrete diffusion model of comparable size. The performance advantage of \methodNAME becomes even more pronounced when compared to autoregressive counterparts such as Nova \cite{nova}, Emu3 \cite{wang2024emu3}, and Lumos-1 \cite{shenoy2024lumosempoweringmultimodal}. While the 8B parameter model, InfinityStar \cite{infinitystar}, currently holds a higher overall score of 83.74, we are confident that the performance gap can be bridged by scaling up the size of \methodNAME. We present qualitative results in Fig. \ref{fig:t2v_examples}. Please refer to Fig.~6 and Fig.~7 in the Appendix for additional results. The generated videos not only accurately capture the semantic details of the user prompts but also maintain a high degree of aesthetic and visual quality.

\subsection{Ablation Studies}

\subsubsection{Predict Indices vs. Predict Bits} 

\begin{wrapfigure}[7]{r}[-0cm]{0.45\textwidth}
	\vspace{-0.4cm}
	\centering
	\captionsetup{font=small, skip=5pt}
	\captionof{table}{Predict Indices vs. Predict Bits on the C2I generation  task.}
	\label{tab:ablation_ind_bit}
	\resizebox{0.4\textwidth}{!}{%
		\begin{tabular}{c|ccc|rr}
			\toprule
			& CFG & interval & $\tau$ & FID$\downarrow$ & IS$\uparrow$ \\
			\midrule
			\methodNAMEcompact$_{ind}$-B & 2.4 & [0.4,1] & 1.33 & 3.56 & 280.3 \\
			\methodNAMEcompact$_{bit}$-B & 2.4 & [0.44,1] & 1.23 & 3.63 & 285.5 \\
			\midline
			\methodNAMEcompact$_{ind}$-L & 2.0 & [0.40,1] & 1.30 & 2.64 & 314.8 \\
			\methodNAMEcompact$_{bit}$-L & 1.9 & [0.45,1] & 1.20 & 2.47 & 287.0 \\
			\bottomrule
		\end{tabular}%
	}
\end{wrapfigure}

As detailed in Sec. \ref{sec:method}, \methodNAME supports predicting either discrete indices (\methodNAMEcompact$_{ind}$) or their binary representations (\methodNAMEcompact$_{bit}$). We compare these two prediction targets on the 256$\times$256 class-conditional image generation task using two model scales: \methodNAME-B (130M) and \methodNAME-L (458M). For each variant, we performed a grid search to identify the optimal decoding parameters, including CFG, CFG interval, and the temperature $\tau$. The results, presented in Tab. \ref{tab:ablation_ind_bit}, indicate that both approaches achieve comparable performance. Specifically, for the smaller \methodNAME-B model, predicting indices yields a slightly better FID score. Conversely, for the larger \methodNAME-L model, predicting bits proves superior, achieving a lower FID of 2.47 compared to 2.64. This suggests that \methodNAME is well-suited for both prediction formats on the class-to-image generation task.

\begin{wrapfigure}[9]{r}{0.65\textwidth}
	\vspace{-0.4cm}
	\centering
	\includegraphics[width=1.0\linewidth]{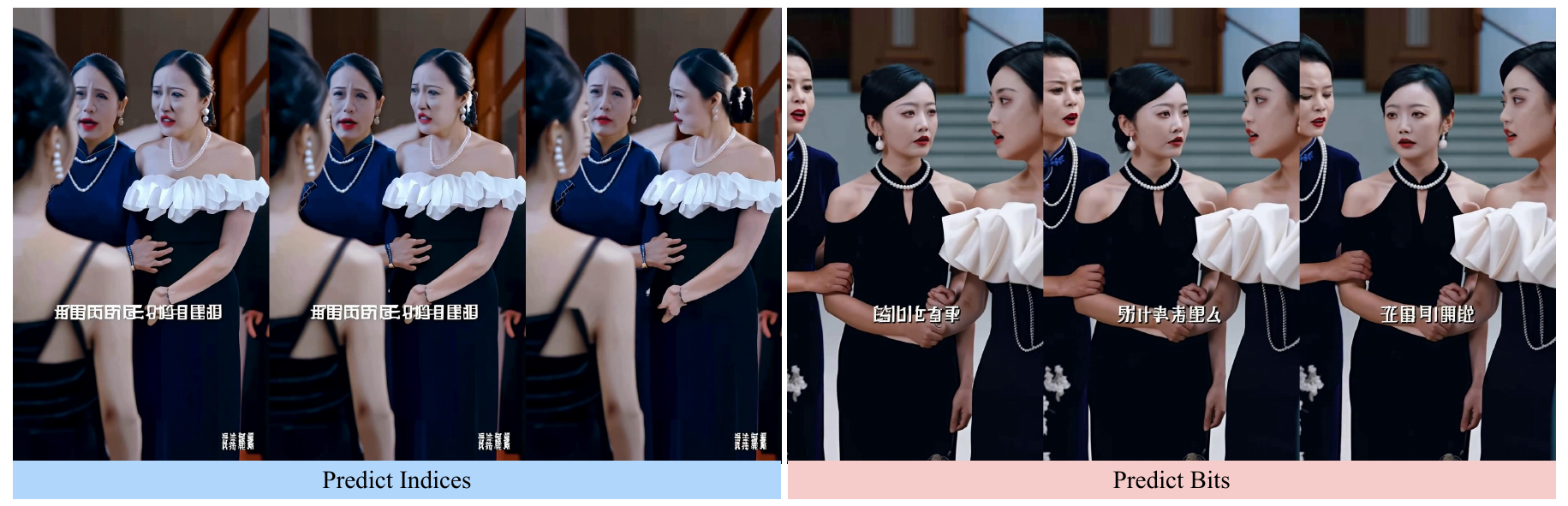}
	\captionsetup{font=small, skip=5pt}
   \vspace{-0.5cm}
	\caption{{Predict Indices vs. Predict Bits on the T2V task.}}
	\label{fig:indices_vs_bits}
\end{wrapfigure}

We further extend this comparison to the more challenging T2V generation task. As illustrated in Fig. \ref{fig:indices_vs_bits}, we observe that the bit prediction approach generates better videos with fewer artifacts. We hypothesize that this is because predicting bits provides a more explicit supervisory signal and mitigates the token aliasing effect inherent in index prediction, thus demonstrating superior performance on complex generation tasks. While some prior works argue that bit prediction assumes independence between bits, leading to suboptimal results, our global refinement mechanism effectively addresses this issue.

\subsubsection{Global Refinement Mechanism}

\begin{wrapfigure}[6]{r}[-1.5em]{0.45\textwidth}
	\vspace{-0.4cm}
	\centering
	\captionsetup{font=small, skip=5pt}
	\captionof{table}{Ablation on Global Refinement Mechanism.}
	\label{tab:ablation_auto_correction}
	\vspace{-0.15cm}
	\resizebox{0.4\textwidth}{!}{
		\begin{tabular}{l|ccc|rr}
			\toprule
			& CFG & interval & $\tau$ & FID$\downarrow$ & IS$\uparrow$ \\
			\midrule
			Refine & 2.4 & [0.44,1] & 1.23 & 3.63 & 285.5 \\
			Mask   & 2.4 & [0.44,1] & 1.23 & 185.62 & 4.3 \\
			Mask   & 8.0 & [0.00,1] & 0.50 & 18.13 & 220.2 \\
			\bottomrule
		\end{tabular}
	}
\end{wrapfigure}

In the ablation study, we validate the effectiveness of our global refinement mechanism, termed Refine. We contrast its performance with a conventional mask-based generation pipeline like MaskGIT \cite{maskgit} or BERT \cite{bert}, where previously generated tokens are fixed. The results are striking: using identical decoding hyperparameters, the mask-based approach collapses into generating nonsensical outputs (FID=185.62), as detailed in Tab. \ref{tab:ablation_auto_correction}. Even with optimal decoding parameters found via grid search (higher CFG, lower temperature $\tau$), the mask-based method (FID=18.13) still lags significantly behind our approach (FID=3.63). This experiment clearly demonstrates that our refinement paradigm effectively mitigates error propagation, a critical weakness in standard AR, and thus achieves superior performance.

\begin{figure}[t]
	\centering
	\includegraphics[width=1.0\linewidth]{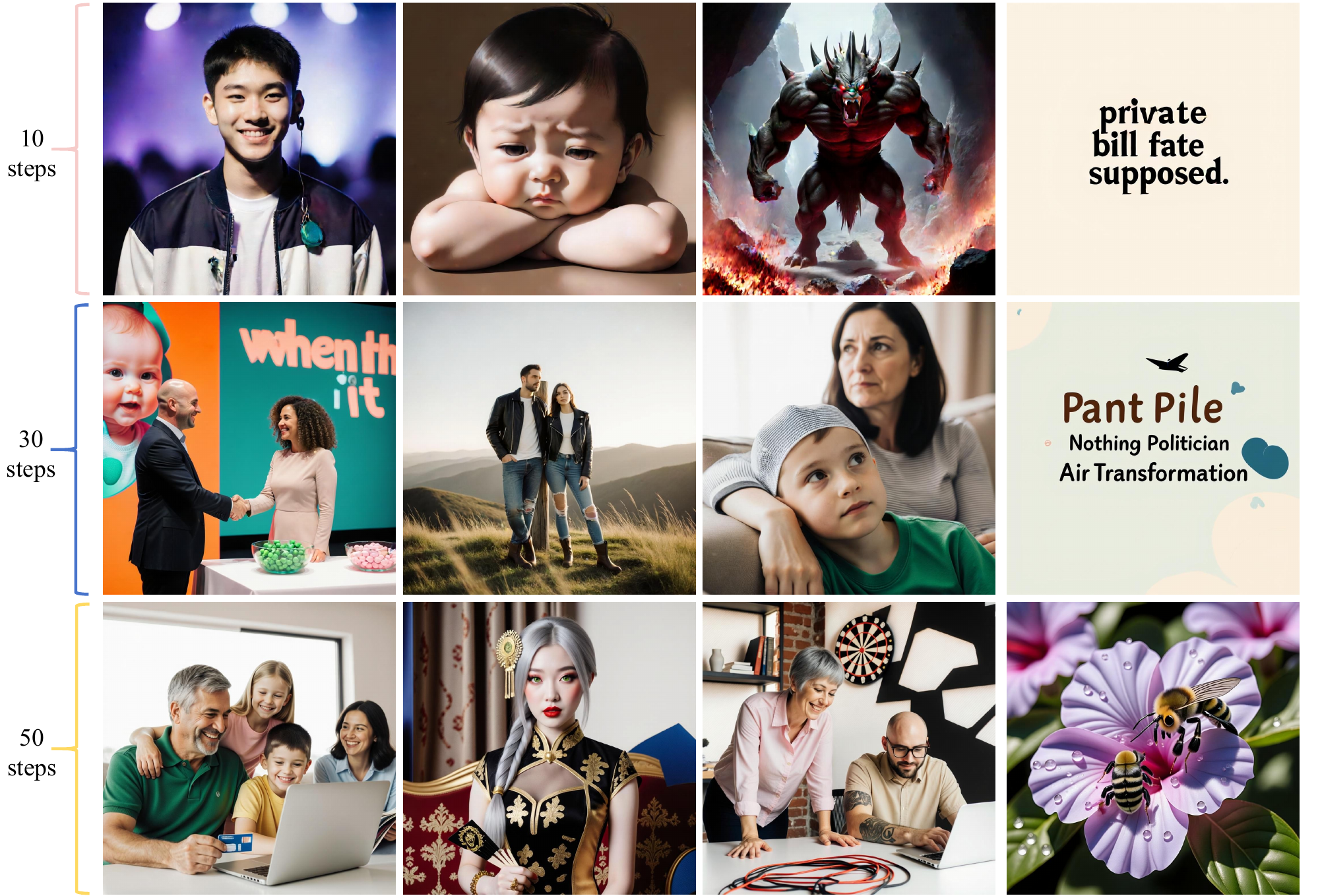}
	\caption{\textbf{Complexity-Aware Sampling: T2I Qualitative Results.}}
	\label{fig:t2i_adaptive_examples}
\end{figure}

\subsubsection{Complexity-Aware Sampling}
\label{sec:complexity_aware_sampling}
\begin{wrapfigure}[11]{r}{0.4\textwidth}
	\vspace{-0.4cm}
	\centering
	\includegraphics[width=1.0\linewidth]{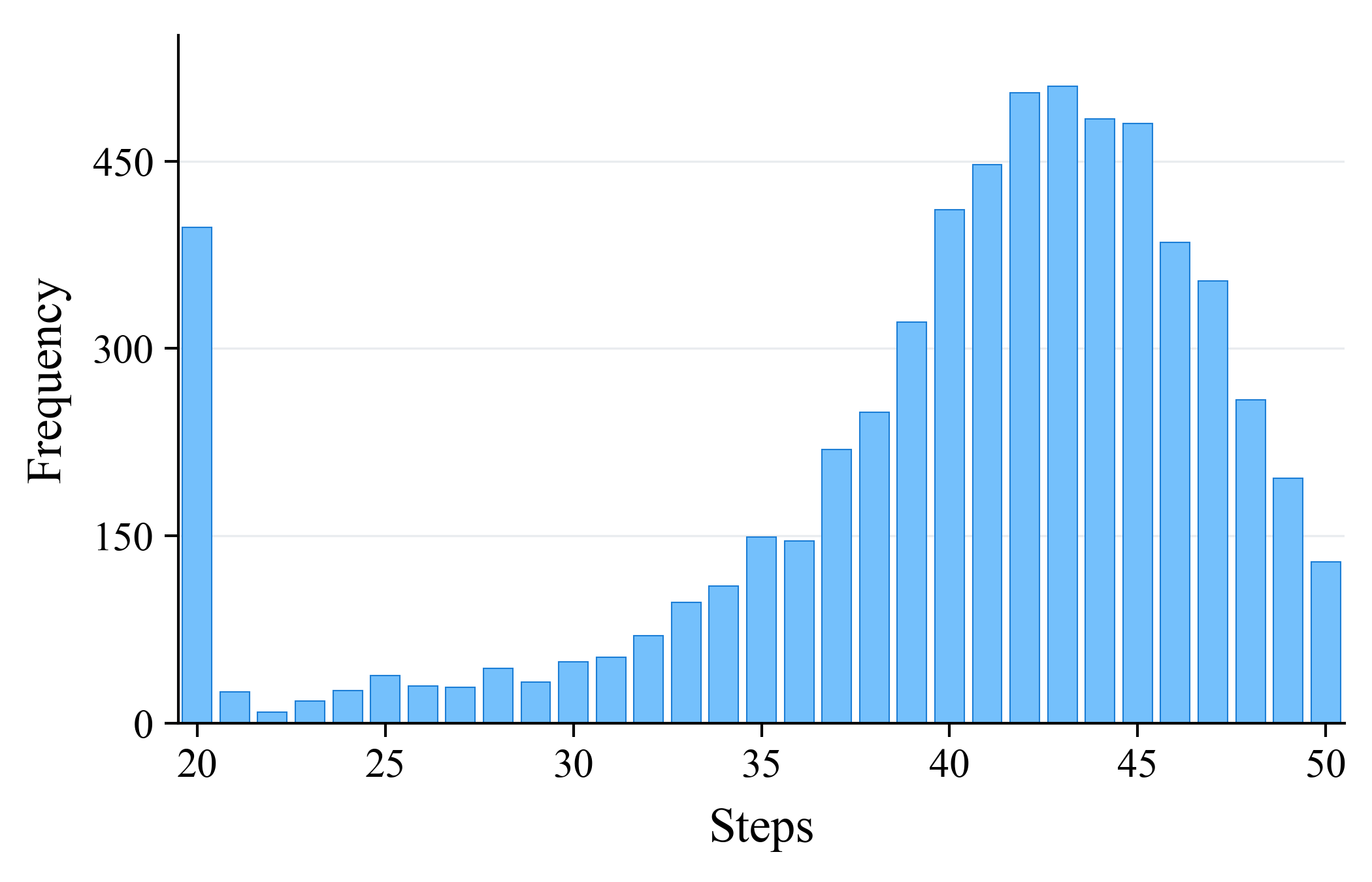}
	\captionsetup{font=small, skip=5pt}
	\caption{{Adaptive-step Generation.}}
	\label{fig:complexity_aware_sampling}
\end{wrapfigure}

We evaluate the efficacy of our complexity-aware sampling on \methodNAMEcompact$_{bit}$-B, with hyperparameters set to $k=700,b=-620$. Following standard settings in diffusion models, we set the maximum number of refinement steps to $T_{max}=50$. To strike a balance between performance and efficiency, we empirically set the minimum number of steps to $T_{min}=20$. We synthesize 63K images and plot the distribution of their allocated generation steps in a histogram (Fig. \ref{fig:complexity_aware_sampling}). The results demonstrate that our proposed method enables \methodNAME to dynamically allocate computational resources based on varying levels of complexity. As observed, different examples are assigned different numbers of refinement steps, ranging from 20 to 50. Over 97.9\% of samples require fewer than 50 refinement steps. Intriguingly, around 6.3\% of images are generated using a minimum of 20 steps, which suggests the model possesses high confidence in these particular predictions. Compared to the fixed-step baseline (50 steps for all samples), our complexity-aware sampling reduces FID from 3.56 to 3.47 while offering significant speedups (max 2.5$\times$, avg 1.25$\times$). Furthermore, we apply complexity-aware sampling to the text-to-image generation task and set $T_{min}=10$. The qualitative results in Fig. \ref{fig:t2i_adaptive_examples} visually confirm that our method effectively enables complexity-aware, adaptive-step generation. For future work, we plan to explore refinement-step distillation, which is naturally compatible with complexity-aware sampling and enables more efficient visual generation.

\subsubsection{Training and Inference Cost}
As shown in Tab.~\ref{tab:in256-sys}, we report Gflops for one forward pass in C2I. Training cost is comparable to one forward pass while inference cost scales with steps. For T2I, \methodNAMEcompact-2B takes 2-8s per 1024$\times$1024 image with an average of 4s (vs. Infinity 1s, FluxDev 12s, Qwen-Image 33s). For T2V, GRN takes 125s for a 480p, 81-frame video (vs. Wan2.1 480s, Wan2.2 437s). Note that these baselines use optimizations from Diffusers, whereas \methodNAME runs in bare PyTorch eager mode. We test these results using the same devices.

We also conduct ablation studies on the bit prediction target for \methodNAMEcompact$_{bit}$ and the decoding parameters. For brevity, please refer to Appendix~\ref{appendix:bit_target} and Appendix~\ref{appendix:decoding_param} for more details.

%% file: sec/5_conclusion.tex
\section{Conclusion}
\label{sec:conclusion}

In this paper, we introduce \methodNAME, a next-generation visual synthesis framework characterized by a global refinement mechanism and complexity-aware generation. We propose Hierarchical Binary Quantization (HBQ) to develop a series of discrete image and video tokenizers that are on par with their continuous counterparts while using the same number of latent channels and offering a significantly higher compression rate. In the generation phase, \methodNAME sets new state-of-the-art results in both image reconstruction and class-conditional image generation. Extensive experiments demonstrate that, at equivalent scales, \methodNAME surpasses existing autoregressive and diffusion-based approaches in both text-to-image and text-to-video generation tasks.

Moreover, as an autoregressive framework built entirely on discrete tokens, we believe \methodNAME could be integrated more naturally into existing large language models. Unified learning over discrete text and visual tokens could substantially promote multimodal understanding and generation. At a fundamental level, \methodNAME resolves the issues of quantization loss and error accumulation that have long limited previous visual autoregressive generative models. We believe it has the potential to emerge as a strong competitor to the currently dominant Transfusion \cite{transfusion} architecture.

\section{Limitations}
This work also has several limitations. Due to limited computational resources, we have not scaled up the training compute or model size to the level of leading visual generation models. In addition, for the text-to-video generation task, we observe that \methodNAME performs better in human-related scenarios. The generated videos may sometimes lack rich visual details and exhibit distortions. We believe that these limitations could be alleviated by balancing the data distribution and scaling up the model size.

\section{Acknowledgements}
We would like to thank Ruibiao Lu for his contributions to data collection and the video demo, and Hui Wu for his valuable advice on infrastructure.

%% file: sec/6_appendix.tex
\section{Algorithm for Hierarchical Binary Quantization}
\label{appendix:hbq_alg}
We outline the procedure for our proposed Hierarchical Binary Quantization (HBQ) in Alg. \ref{alg:hbq}. The HBQ binary tokens are ordered from coarse to fine. Early tokens in the sequence correspond to core semantic concepts, while later tokens introduce high-frequency details, progressively enriching the representation. Furthermore, our multi-round quantization process generates binary tokens sequentially, with each round corresponding to a single bit. This inherent structure makes HBQ particularly well-suited for direct, bitwise prediction tasks. In contrast to methods like FSQ \cite{fsq}, which quantize vectors holistically, our approach offers a more natural framework for bitwise generation, \emph{i.e.}, \methodNAMEcompact$_{bit}$.

\begin{algorithm}[h]
	\caption{Hierarchical Binary Quantization} \label{alg:hbq}
	\begin{algorithmic}[0]
		\Require VAE encoder $\mathcal{E}$, VAE decoder $\mathcal{D}$, an image or a video $\bm{X}$,  HBQ round $M$,
		$\delta(\cdot)$ function where $\delta(0)=-1$ and $\delta(1)=1$
		\State $\bm{Q}_{queue} \gets []$ \Comment{HBQ binary labels} \\
		$\bm{F} = \mathcal{E}(\bm{X})$ \Comment{VAE Encoding}  \\
		$\bm{F} = tanh \left( \bm{F} \right)$ \Comment{restrict data range to [-1,1]}  \\
		$\bm{c_1}=\operatorname{ZerosLike}(\bm{F})$ \Comment{initialize bucket centroids}  
		\For {$i=1,2,\ldots,M$}   \Comment{HBQ rounds}
		\State $\bm{q_i}=(\bm{F}>\bm{c_i})$ \Comment{binary quantization}  
		\State $\operatorname{QueuePush}(\bm{Q}_{queue}, \bm{q_i})$ \Comment{update HBQ binary labels}  
		\State $\bm{c_{i+1}}=\bm{c_i}+ \delta[\bm{q_i}] \cdot 2^{-i} $ \Comment{update bucket centroids for next round}  
		\EndFor \\
		$\bm{\hat{F}}=\delta[\bm{q_1}] \cdot 2^{-1} + \delta[\bm{q_2}] \cdot 2^{-2} + ...+\delta[\bm{q_M}] \cdot 2^{-M}$  \Comment{obtain quantized feature}  \\
		$\bm{\hat{F}}=\operatorname{StopGrad}[\bm{\hat{F}}-\bm{{F}}]+\bm{{F}}$ \Comment{Straight-Through Estimator}\\
		$\bm{\hat{X}}=\mathcal{D}(\bm{\hat{F}})$  \Comment{VAE Decoding}
		
		\Ensure $\bm{Q}_{queue}=[\bm{q_1},\bm{q_2},...,\bm{q_M}]$, $\bm{\hat{F}}$, $\bm{\hat{X}}$
	\end{algorithmic}
\end{algorithm}

\section{Algorithm for Generative Refinement Network}
\label{appendix:grn_alg}
Alg. \ref{alg:grn_train} outlines the pseudo-code for a single training step. The model input, denoted as $F_t$, is a hybrid feature map composed of a subset of ground truth tokens and a complementary subset of random tokens. Taking $F_t$ as input, \methodNAME is trained to predict the complete set of ground truth tokens. Despite its simplicity, this training strategy implicitly teaches the model to differentiate between reliable (ground truth) and unreliable (random) ones. Consequently, the model learns to preserve reliable tokens while refining the unreliable ones.

\definecolor{codeblue}{rgb}{0.25,0.5,0.5}
\definecolor{codekw}{rgb}{0.85, 0.18, 0.50}

\definecolor{codesign}{RGB}{0, 0, 255}
\definecolor{codefunc}{rgb}{0.85, 0.18, 0.50}

\lstdefinelanguage{PythonFuncColor}{
	language=Python,
	keywordstyle=\color{blue}\bfseries,
	commentstyle=\color{codeblue},
	stringstyle=\color{orange},
	showstringspaces=false,
	basicstyle=\ttfamily\small,
	literate=
	{*}{{\color{codesign}* }}{1}
	{-}{{\color{codesign}- }}{1}
	{+}{{\color{codesign}+ }}{1}
	{/}{{\color{codesign}/ }}{1}
	{dataloader}{{\color{codefunc}dataloader}}{1}
	{sample_t}{{\color{codefunc}sample\_t}}{1}
	{randn}{{\color{codefunc}randn}}{1}
	{randn_like}{{\color{codefunc}randn\_like}}{1}
	{jvp}{{\color{codefunc}jvp}}{1}
	{stopgrad}{{\color{codefunc}stopgrad}}{1}
	{l2_loss}{{\color{codefunc}l2\_loss}}{1}
	{net_fn}{{\color{codefunc}net}}{1}
	{cross_entrophy_loss}{{\color{codefunc}cross\_entrophy\_loss}}{1}
}

\lstset{
	language=PythonFuncColor,
	backgroundcolor=\color{white},
	basicstyle=\fontsize{8pt}{8.4pt}\ttfamily\selectfont,
	columns=fullflexible,
	breaklines=true,
	captionpos=b,
}

The sampling procedure of \methodNAME is detailed in Alg. \ref{alg:grn_sample} (we omit complexity-aware sampling for clarity). The process is analogous to human drawing, where a state is iteratively refined. At each step $t$, the current state $F_t$ consists of the already drawn content, represented by $S_t \cdot Y_t$, and the remaining blank regions filled with random tokens, represented by $\overline{S_t} \cdot Y_{rand}$. The model then predicts a complete set of tokens based on the current state. A randomly selected subset of these new predictions is then used to update the state for the next refinement step. This straightforward random selection mechanism elegantly unifies three essential operations into a single framework:

\begin{itemize}
	\item Filling: Introducing predicted tokens into previously blank areas.
	\item Refining: Improving the quality of previously predicted tokens.
	\item Erasing: Replacing previously predicted tokens with random ones.
\end{itemize}

To better illustrate this process, both Fig. \ref{fig:framework} in the paper and the fourth column of Fig. \ref{fig:ar_compare} in the appendix visualize the sampling process of \methodNAME. We especially highlight concrete examples of the filling, refining, and erasing tokens described above.

\begin{algorithm}[h]
\caption{Generative Refinement Networks: Training Step}
\label{alg:grn_train}
\begin{lstlisting}
# net(z): GRN network
# y_gt: HBQ index labels or HBQ bit labels
# C: classes, C=2^n for index labels, C=2 for bit labels

pt = sample_pt()
y_rand = randint(C, y_gt.shape)

st = rand_like(y_gt) < pt
ft = st * y_gt + logical_not(st) * y_rand

y_pred = net(ft)
loss = cross_entropy_loss(y_pred, y_gt)  
\end{lstlisting}
\end{algorithm}

\begin{algorithm}[h]
\caption{Generative Refinement Networks: Sampling Process}
\label{alg:grn_sample}
\begin{lstlisting}
# We omit Complexity Aware Sampling for ease of understanding

y_rand = randint(C, target_shape)
y_pred = y_rand
for t in range(T):
pt = (t + 1) / T
st = rand_like(y_pred) < pt
ft = st * y_pred + logical_not(st) * y_rand
y_pred = net(ft)
\end{lstlisting}
\end{algorithm}

\begin{figure}[t]
	\centering
	\includegraphics[width=1.0\linewidth]{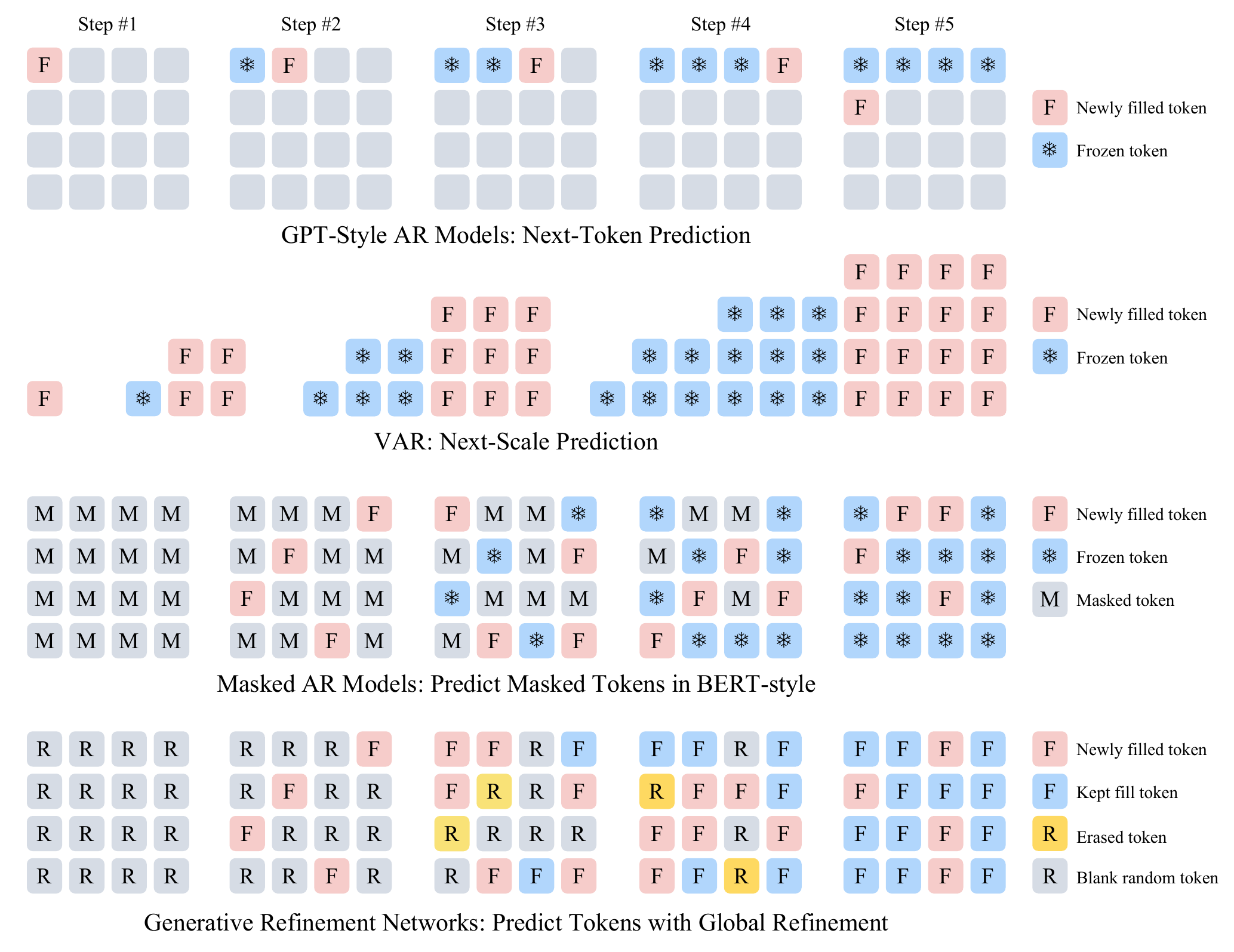}
	\caption{\textbf{Comparison between \methodNAME with other autoregressive models in visual generation.} With the global refinement mechanism, \methodNAME iteratively revises and enhances the entire visual representation, effectively mitigating the error propagation issue in conventional autoregressive models.}
	\label{fig:ar_compare}
\end{figure}

\section{Difference with Other Autoregressive Models}
\label{sec:appendix_ar_diff}

In Fig. \ref{fig:ar_compare}, we compare \methodNAME with conventional autoregressive models, including GPT-Style AR models (next-token prediction) \cite{llamagen}, VAR (next-scale prediction) \cite{keyuVAR}, and Masked AR models (BERT-Style) \cite{maskgit}. Conventional AR models are constrained by a fixed generation order, where previously generated tokens are immutable. This often leads to issues like error propagation. In stark contrast, \methodNAME employs a flexible global refinement strategy, leveraging its unique filling, refining, and erasing mechanism. This allows our model to iteratively revise and enhance the entire visual representation, effectively mitigating the error propagation issue inherent in conventional autoregressive models.

\section{Implementation Details}
\label{appendix:implementation}
\textbf{Model Architecture.} Tab. \ref{tab:model_detail} summarizes the architectural details of our proposed C2I, T2I, and T2V models. Following the methodology of JiT \cite{jitpaper}, we implement four variants for the C2I task, with model sizes scaling from 130M to 2B parameters. For T2I and T2V generation, we introduce a new 2B-parameter architecture specifically designed to meet the FlexAttention requirement, which requires the head dimension to be a multiple of 128. Furthermore, our models support sequence packing to accelerate training, and NaViT \cite{navit} to handle arbitrary aspect ratios and resolutions.

\textbf{Visual Tokenizer.} For the C2I generation task, we employ an image-only visual tokenizer trained on the OpenImages dataset \cite{openimages}. This tokenizer features a latent dimension of 16 and utilizes 4 rounds of HBQ. It compresses a 256 $\times$ 256 image into  16 $\times$ 16 $\times$ 16 $\times$ 4 binary tokens with a spatial stride of 16, achieving a state-of-the-art reconstruction FID of 0.56 on the ImageNet benchmark. In contrast, our T2I and T2V models share a unified visual tokenizer, which is jointly optimized on a mixture of images and videos. This unified tokenizer is configured with a 64-dimensional latent space, a spatial stride of 16, a temporal stride of 4, and also undergoes 4 rounds of HBQ, attaining a FVD of 26.3 and a PSNR of 34.73 on our video reconstruction benchmark. Additional results for the released tokenizers are provided in Tab. \ref{tab:imagenet_recons} and Tab. \ref{tab:video_recons} in the paper, as well as Tab. \ref{tab:gan_loss} in the appendix.

\textbf{Training.} For our C2I models (\methodNAME-B/L/H/G), we conduct training on the ImageNet dataset using 256×256 resolution images. The models are trained for 600 epochs, equivalent to 750K iterations, with a batch size of 1024. We employ a constant learning rate of 2e-4 throughout the training process. We apply a 10\% condition dropping rate to enable Classifier-Free Guidance. For our \methodNAME-T2I and \methodNAME-T2V models, we adopt a coarse-to-fine training strategy. For instance, the \methodNAME-T2I model is trained for 150K iterations at 256$\times$256 resolution and 60K iterations at 1024$\times$1024 resolution with batch sizes of 15400 and 2048, respectively, using corresponding learning rates of 2e-4 and 2e-5. Other hyperparameters, such as the constant learning rate schedule, zero weight decay, and 10\% condition drop, remain consistent across these models.

\textbf{Sampling.} During the inference phase, we utilize Classifier-Free Guidance to enhance sample quality and adherence to conditioning. For our ImageNet-trained models, we dynamically start CFG based on an optimal threshold $p_t$ found within the range of [0, 0.5]. The CFG strength is swept within [1.0, 3.0] to find the best results. For our text-conditional models (\methodNAME-T2I and \methodNAME-T2V), we apply CFG throughout the entire sampling process. We search for the optimal CFG strength within the range of [1.0, 4.0] and a temperature $\tau$ range of [0.5, 1.5] to generate diverse and high-fidelity results. For the benchmark experiments on the C2I, T2I, and T2V tasks, we use 50 fixed refinement steps. In Sec.~\ref{sec:complexity_aware_sampling}, we apply complexity-aware sampling to the C2I and T2I tasks. The proposed complexity-aware sampling method is controlled by parameters $k$ and $b$, where $k$ is optimized in the range of [300, 1200] and $b$ is set accordingly to determine overall sampling steps. 

\begin{table}[ht]
	\centering
	\vspace{-0cm}
	\caption{\textbf{Implementation details for our C2I, T2I, T2V models.} We elaborate on four key aspects: model architecture, visual tokenizer, training, and sampling.}
	\vspace{-0cm}
	\label{tab:model_detail}
	\begin{tabular}{lcccccc}
		\toprule
		& \textbf{\methodNAME-B} & \textbf{\methodNAME-L} & \textbf{\methodNAME-H} & \textbf{\methodNAME-G} & \textbf{\methodNAME-T2I} & \textbf{\methodNAME-T2V} \\
		\midrule
		\rowcolor{lightgray}
		\multicolumn{7}{l}{\textbf{Architecture}} \\
		Depth & 12 & 24 & 32 & 40 & 28 & 28\\
		Hidden dim & 768 & 1024 & 1280 & 1664 & 2304 & 2304 \\
		Heads & 12 & 16 & 16 & 16 & 18 & 18 \\
		Parameter & 130M & 458M & 952M & 2B & 2B & 2B \\
		Data & \multicolumn{4}{c}{ImageNet 256$\times$256 images} & Images & Videos \\
		\rowcolor{lightgray}
		\multicolumn{7}{l}{\textbf{Tokenizer}} \\
		Tokenizer Stride & \multicolumn{4}{c}{16$\times$16} & \multicolumn{2}{c}{16$\times$16$\times$4} \\
		Tokenizer dim & \multicolumn{4}{c}{16} & \multicolumn{2}{c}{64} \\
		HBQ Round & \multicolumn{4}{c}{4} & \multicolumn{2}{c}{4} \\
		\rowcolor{lightgray}
		\multicolumn{7}{l}{\textbf{Training}} \\
		Epoch & \multicolumn{4}{c}{600} & \multicolumn{2}{c}{N/A} \\
		Iter & \multicolumn{4}{c}{750K} & 150K/60K & 150K/9K \\
		Batch Size & \multicolumn{4}{c}{1024} & 15400/2048 & 4096/1350 \\
		LR & \multicolumn{4}{c}{2e-4} & 2e-4/2e-5 & 2e-4/2e-5 \\
		LR Schedule & \multicolumn{4}{c}{constant} & \multicolumn{2}{c}{constant} \\
		Weight Decay & \multicolumn{4}{c}{0} & \multicolumn{2}{c}{0} \\
		Condition Drop & \multicolumn{4}{c}{10\%} & \multicolumn{2}{c}{10\%} \\
		\rowcolor{lightgray}
		\multicolumn{7}{l}{\textbf{Sampling}} \\
		$k$,$b$ &  \multicolumn{6}{c}{optimal $k$ in [300,1200], $b$ is set accordingly}  \\
		$\tau$ sweep range & \multicolumn{4}{c}{[1.0,1.5]} & \multicolumn{2}{c}{[0.5,1.5]} \\
		CFG sweep range & \multicolumn{4}{c}{[1.0, 3.0]} & \multicolumn{2}{c}{[1.0,4.0]} \\
		CFG interval & \multicolumn{4}{c}{optimal $p_t$ in [0, 0.5] to start CFG } & \multicolumn{2}{c}{not used} \\
		
		\bottomrule
	\end{tabular}
	\vspace{-0cm}
	\label{tab:architectures}
\end{table}

\section{More Ablation Studies}
\subsection{GAN Loss in Tokenizer}
\label{appendix:tune_gan}

Fig. \ref{fig:ablation_hb_round} illustrates a clear trend: reconstruction quality steadily improves as more HBQ rounds are introduced. Although four to six rounds are sufficient for decent performance, an eight-round model closes the gap, achieving results nearly the same as the baseline without quantization.

We further investigate the impact of varying the GAN loss weight, $\lambda_{GAN}$. For this study, we fine-tune a baseline model (an HBQ tokenizer with 64 latent channels and four quantization rounds) using different $\lambda_{GAN}$ values. As detailed in Tab. \ref{tab:gan_loss}, increasing the GAN loss weight from 0.001 to 0.02 significantly improves perceptual quality, reducing the rFVD from 48.6 to 28.6. However, this comes at the cost of a drop in reconstruction fidelity, with the PSNR decreasing from 34.05 to 33.73. Empirically, we found that a weight of 0.005 strikes an effective balance between these two metrics. Consequently, this setting is applied to our \methodNAMEcompact-T2I and \methodNAMEcompact-T2V models.

\begin{table*}[t]
	\centering
	\caption{\textbf{Comparison between different GAN loss weights.}}
	\label{tab:gan_loss}
	\resizebox{1.0\linewidth}{!}{ 
		\begin{tabular}{l|c|ccc|rccc}
			\toprule
			Method &  $\lambda_{GAN}$ & Channel & Stride & Compress  & {rFVD$\downarrow$} & {LPIPS$\downarrow$} & {SSIM$\uparrow$} & {PSNR$\uparrow$} \\
			\midrule
			Wan 2.1 (patchify) & N/A & 64 & 16$\times$16$\times$4 & 24 & 19.5  & 0.058 & 0.929 & 34.10 \\
			\midline
			HBQ (baseline) & 0.001 & 64 & 16$\times$16$\times$4  & 96 & 50.6  & 0.084 & 0.930 & 33.97 \\
			\midline
			HBQ ($M=4$) & 0.001 & 64 & 16$\times$16$\times$4  & 96 & 48.6  & 0.083 & \textbf{0.930} & \textbf{34.05} \\
			HBQ ($M=4$) & 0.005 & 64 & 16$\times$16$\times$4  & 96 & 30.0  & \textbf{0.078} & 0.928 & 33.98 \\
			HBQ ($M=4$) & 0.02 & 64 & 16$\times$16$\times$4  & 96 & \textbf{28.6}  & 0.081 & 0.925 & 33.73 \\
			HBQ ($M=4$) & 0.1 & 64 & 16$\times$16$\times$4  & 96 & 29.5  & 0.084 & 0.923 & 33.55 \\
			\bottomrule
			
		\end{tabular}
	}
\end{table*}

\begin{wrapfigure}[8]{r}{0.4\textwidth} 
	\centering
	\vspace{-0.4cm}
	\captionsetup{font=small, skip=5pt}
	\captionof{table}{Impacts of HBQ rounds on generation performance. Here we test \methodNAMEcompact$_{bit}$-B with tokenizers of different HBQ rounds}
	\vspace{-0.05cm}
	\label{tab:hbq_round}
	\resizebox{\linewidth}{!}{ 
		\begin{tabular}{c|cc|rr|rr}
			\toprule
			& \multicolumn{2}{c|}{Recons} & \multicolumn{2}{c|}{GRN-B(130M)} & \multicolumn{2}{c}{GRN-L(458M)} \\
			& rFID$\downarrow$ & PSNR$\uparrow$ & gFID$\downarrow$ & IS$\uparrow$ & gFID$\downarrow$ & IS$\uparrow$ \\
			\midrule
			HBQ-2 & 1.23 &21.24 & 3.22 & 304.1 & 2.71 & 309.4 \\
			HBQ-4 & 0.54 & 22.81 & 3.54 & 293.0 & 2.63 & 290.1 \\
			HBQ-6 & 0.49 & 23.06  & 4.74 & 235.2 & 3.10 & 253.6 \\
			HBQ-8 & 0.48 & 23.09  & 5.29 & 208.8 & 3.48 & 235.2 \\
			\bottomrule
		\end{tabular}
	}
\end{wrapfigure}

	\subsection{Impacts of HBQ Rounds on Generation Performance}
	\label{appendix:hbq_round}
	
	As shown in Tab.~\ref{tab:hbq_round}, although HBQ (M=2) performs best for \methodNAMEcompact-B with 130M parameters. HBQ (M=4) surpasses HBQ (M=2) when scaled to \methodNAME-L (458M parameters). We conclude that large models benefit from more HBQ rounds. Another interesting observation is that although the reconstruction metrics remain nearly identical from 6 rounds to 8 rounds, there are notable discrepancies in the generation metrics. This indicates that larger model capacity is indeed required when information is distributed across more distinct bits.
	Additionally, our experiments reveal that the accuracy of MSBs is substantially higher than that of LSBs. Future comparative studies between HBQ (16dim, M=8) and HBQ (32dim, M=4) would constitute a compelling line of experimentation.

\subsection{Binary Selection Map: Random vs. Confidence}
\label{appendix:random_confidence}

\begin{wrapfigure}[6]{r}[-0cm]{0.45\textwidth}
	
	\centering
	\vspace{-0.4cm}
	\captionsetup{font=small, skip=5pt}
	\captionof{table}{Random sampling vs. Confidence sampling on the C2I generation task.}
	\vspace{-0.15cm}
	\label{tab:random_confidence_sample}
	\resizebox{0.4\textwidth}{!}{
		\begin{tabular}{c|cc|rr}
			\toprule
			& \multicolumn{2}{c|}{Sampling} & \multicolumn{2}{c}{Metrics} \\
			& Random & Confidence & FID$\downarrow$ & IS$\uparrow$ \\
			\midrule
			\methodNAMEcompact$_{bit}$-B & \checkmark &  & 3.63 & 285.5 \\
			\methodNAMEcompact$_{bit}$-B &  & \checkmark & 10.64 & 246.8 \\
			\bottomrule
		\end{tabular}
	}
\end{wrapfigure}
As detailed in Eq. \ref{eq:mt_pt}, the binary selection map $S_t$ is constructed without prior constraints during generation, meaning that we randomly select current predictions to update the state $F_{t+1}$ for the next refinement step. To understand the importance of this random sampling, we conducted an experiment using a confidence-based sampling alternative. In this setting, tokens to update $F_{t+1}$ are selected based on their prediction confidence. While this prioritizes tokens deemed more `correct', the outcome was a severe performance drop (FID: 3.63 $\rightarrow$ 10.64), as detailed in Tab. \ref{tab:random_confidence_sample}. The reason for this counter-intuitive result lies in the discrepancy between training and inference patterns. Our model is trained to operate on a state where ground truth and random tokens are uniformly distributed. The confidence-based method breaks this assumption by selecting high-confidence tokens that are not uniformly distributed but are instead clustered. This distributional shift moves the input far from the manifold learned during training, resulting in a catastrophic failure of the generative process.

\subsection{Bit Prediction Target: Absolute vs. Relative}
\label{appendix:bit_target}
\begin{figure}[h]
	\centering
	\includegraphics[width=1.0\linewidth]{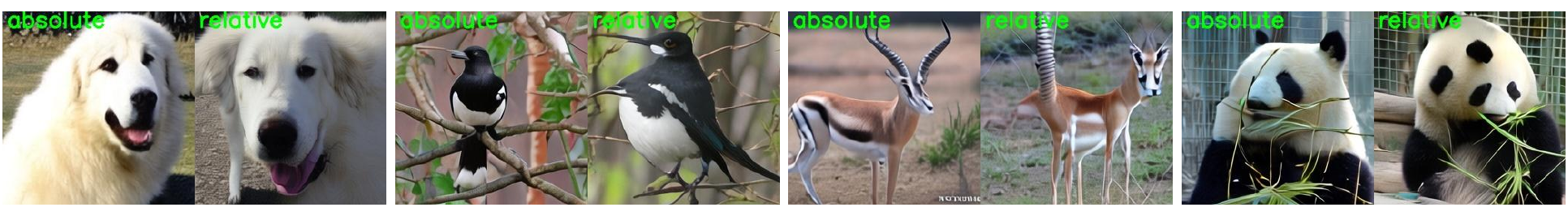}
	\caption{\textbf{Comparison of Absolute and Relative Bit Prediction.}}
	\vspace{-0.6cm}
	\label{fig:absolute_relative}
\end{figure}

We investigate two different prediction targets for the binary labels in \methodNAMEcompact$_{bit}$: absolute bits versus relative bits. While absolute bit prediction directly targets the ground-truth bits ($Y_{gt}$), relative bit prediction targets whether the input bit should be flipped. This can be formulated as predicting a residual, $Y^{rel}_{gt}= (F_{t} \neq Y_{gt})$, where a `1' indicates a required flip and a `0' indicates preservation. As shown in Fig. \ref{fig:absolute_relative}, our experiments reveal that predicting absolute bits yields superior results, leading to generated images with significantly better structural stability compared to those from relative bit prediction.

	\subsection{Decoding Hyper-Parameters}
	\label{appendix:decoding_param}

	\begin{wrapfigure}[11]{r}{0.5\textwidth} 
		\centering
		\vspace{-0.4cm}
		\captionsetup{font=small, skip=5pt}
		\captionof{table}{Comparisons of different sampling methods and sensitivity analysis for complexity-aware sampling.}
		\vspace{-0.0cm}
		\label{tab:sensitive}
		
		\resizebox{0.9\linewidth}{!}{ 
			\begin{tabular}{c|cc|cc|rr|rr}
				\toprule
				Method & $k$ & $b$ & $T_{min} $ & $T_{max}$ & $T_{avg}$ & $T_{std}$ & FID$\downarrow$ & IS$\uparrow$ \\
				\midrule
				Linear & / & / & 50 & 50 & 50 & 0.0 & 3.56 & 290.3 \\
				Cosine & / & / & 50 & 50 & 50 & 0.0 & 3.94 & 272.5 \\
				Complexity  & 700 & -620 & 20 & 50 & 40 & 7.4 & 3.47 & 287.3 \\
				\midrule
				Linear & / & / & 30 & 30 & 30 & 0.0 & 5.06 & 262.3 \\
				Cosine & / & / & 30 & 30 & 30 & 0.0 & 7.87 & 224.5 \\
				Complexity  & 700 & -630 & 20 & 40 & 24  & 4.0 & 3.79 & 270.9 \\
				\midrule
				\multirow{5}{*}{Complexity} & 600 & -527 & 20 & 50 & 39 & 6.8 & 3.50 & 286.6 \\
				& 600 & -527 & 10 & 50 & 39 & 8.2 & 3.53 & 282.1 \\
				& 700 & -620 & 20 & 50 & 40 & 7.4 & 3.47 & 287.3 \\
				& 700 & -630 & 20 & 40 & 24  & 4.0 & 3.79 & 270.9 \\
				& 800 & -714 & 20 & 50 & 40 & 7.9 & 3.47 & 285.9 \\
				\bottomrule
			\end{tabular}
		}
	\end{wrapfigure}

\begin{figure}[h]
	\centering
	\includegraphics[width=1.0\linewidth]{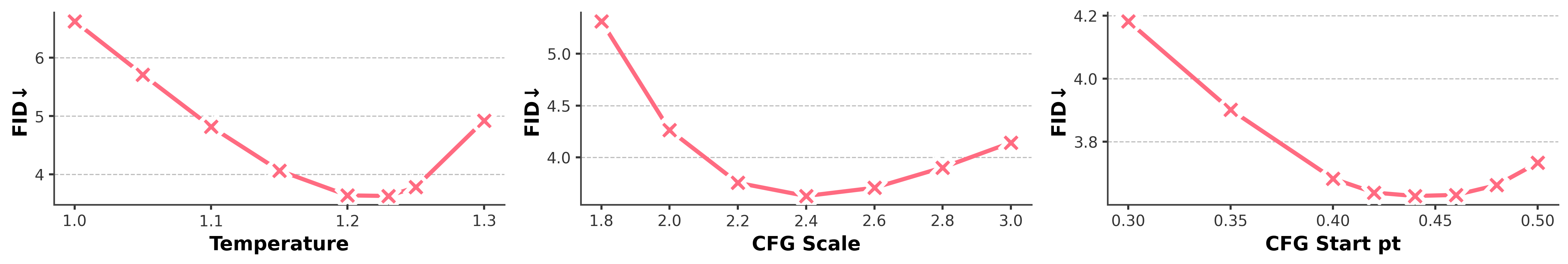}
	\caption{\textbf{Influence of decoding hyper-parameters: $\tau$, CFG, and CFG start $p_t$.}}
	\label{fig:decoding_param}
\end{figure}

	Regarding the decoding parameters, we use \methodNAMEcompact$_{bit}$-B as a representative example. The optimal parameters were found to be $\tau=1.23$, $CFG=2.4$, and an interval of $[0.44,1]$. We empirically observed that a lot of parameter combinations can achieve similar results. Intuitively, increasing $\tau$ or decreasing $CFG$ encourages more diverse generation but can also lead to greater instability. The CFG interval is introduced to restore diversity when applying a higher CFG strength, as it disables CFG during the initial decoding steps, which are crucial for determining the overall semantics. The effect of varying each parameter while keeping the others fixed is detailed in Fig. \ref{fig:decoding_param}.  As shown in Tab. \ref{tab:sensitive}, our complexity-aware sampling outperforms linear and cosine schedules in fewer steps. In our complexity-aware sampling, the main parameters are $k$ and $b$. We tune $k$ for dynamic range and $b$ for average steps. $T_{min}$ and $T_{max}$ clip the number of steps to a reasonable range and require less tuning. Tab. \ref{tab:sensitive} shows good results with suitable ranges for $k$ in [600, 800].

\section{More Qualitative Results}
\subsection{C2I Qualitative Results}

\begin{figure}[t]
	\centering
	\includegraphics[width=1.0\linewidth]{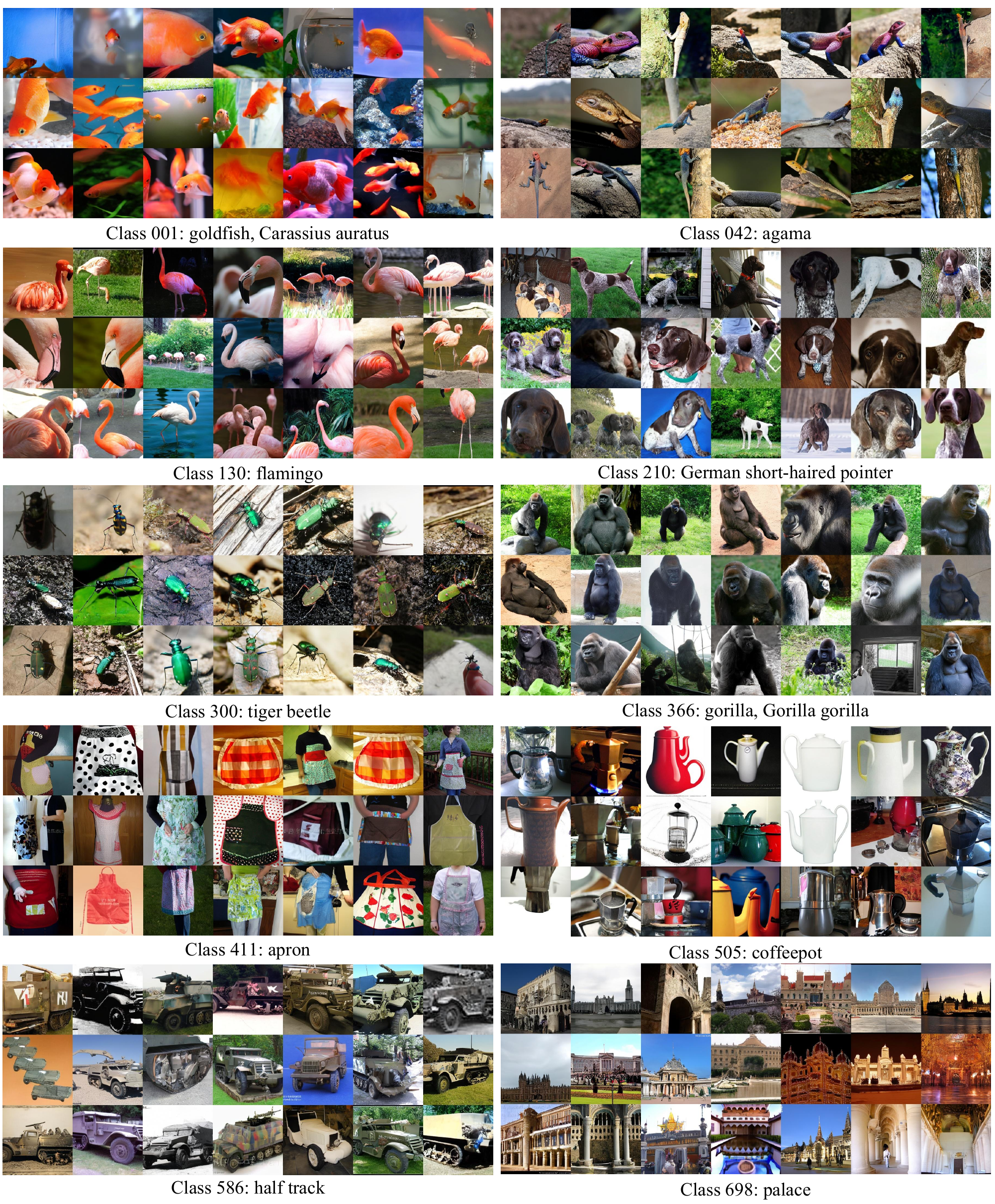}
	\vspace{-0.3cm}
	\caption{\textbf{Uncurated 256$\times$256 samples from \methodNAME-G on ImageNet}. To ensure representative results, these images are generated using the same parameters that yielded our reported FID of 1.81 (CFG scale = 1.7, CFG interval = [0.3, 1.0]), rather than using a higher CFG scale typically favored for visualization.}
	\label{fig:c2i_uncurated}
\end{figure}

Similar to JiT \cite{jitpaper}, we present uncurated 256$\times$256 samples generated by \methodNAME-G in Fig. \ref{fig:c2i_uncurated}. To ensure a fair and representative visualization of our model's capabilities, these images were generated with the same CFG scale 1.7 and  CFG interval = [0.3, 1.0] used to achieve the best FID of 1.81. This contrasts with the common approach of using a higher CFG scale for qualitative examples, which may not reflect the model's real performance as measured by FID.

\subsection{T2I Qualitative Results}
\begin{figure}[h!]
	\centering
	\vspace{-0.9cm}
	\includegraphics[width=1.0 \linewidth]{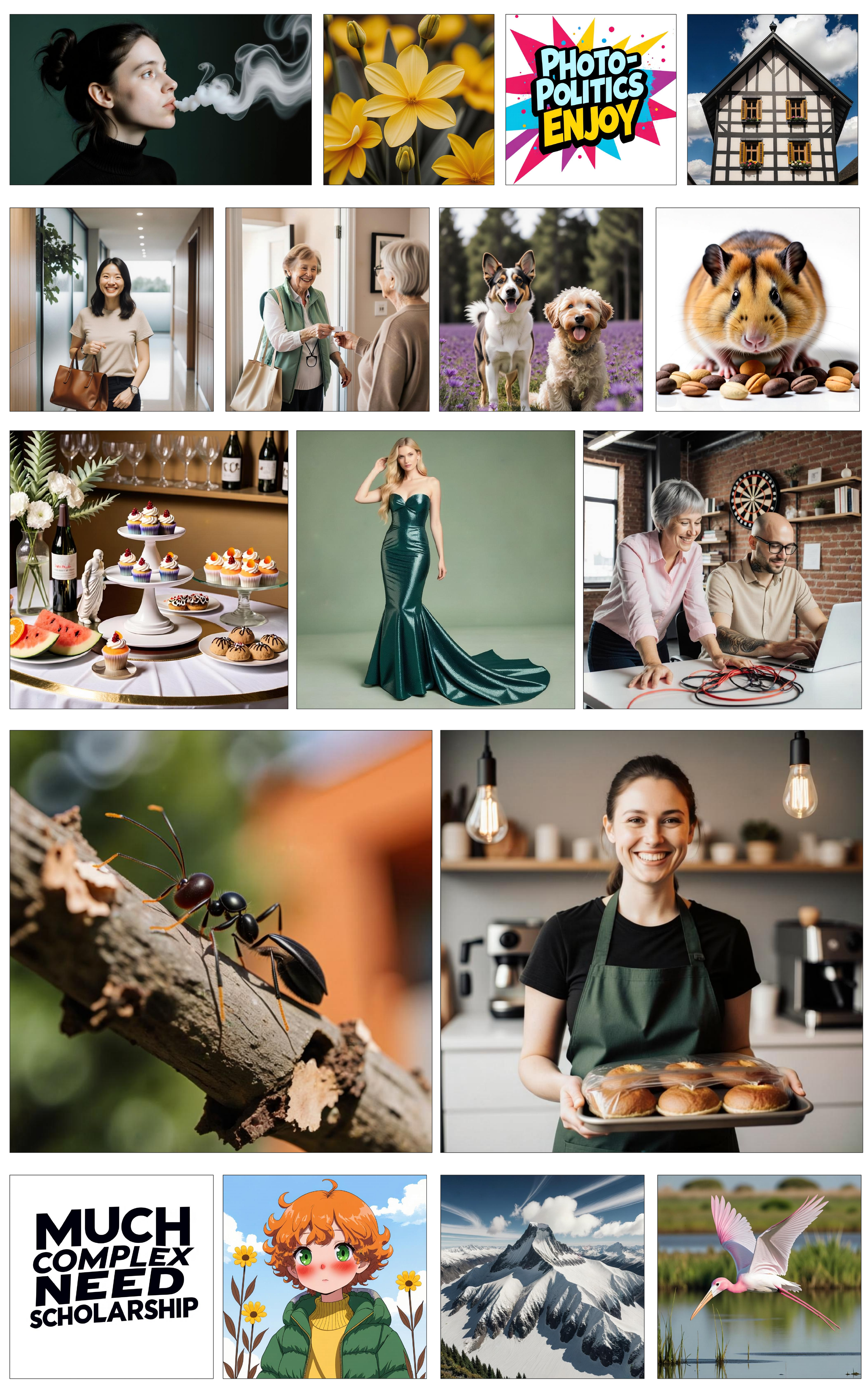}
	\caption{\textbf{More qualitative results for the text-to-image generation task.}}
	\label{fig:t2i_examples}
\end{figure}

In Fig. \ref{fig:t2i_examples}, we present 1024$\times$1024 images generated by \methodNAME-T2I.

\subsection{T2V Qualitative Results}
\begin{figure}[h!]
	\centering
	\vspace{-0.7cm}
	\includegraphics[width=1.0\linewidth]{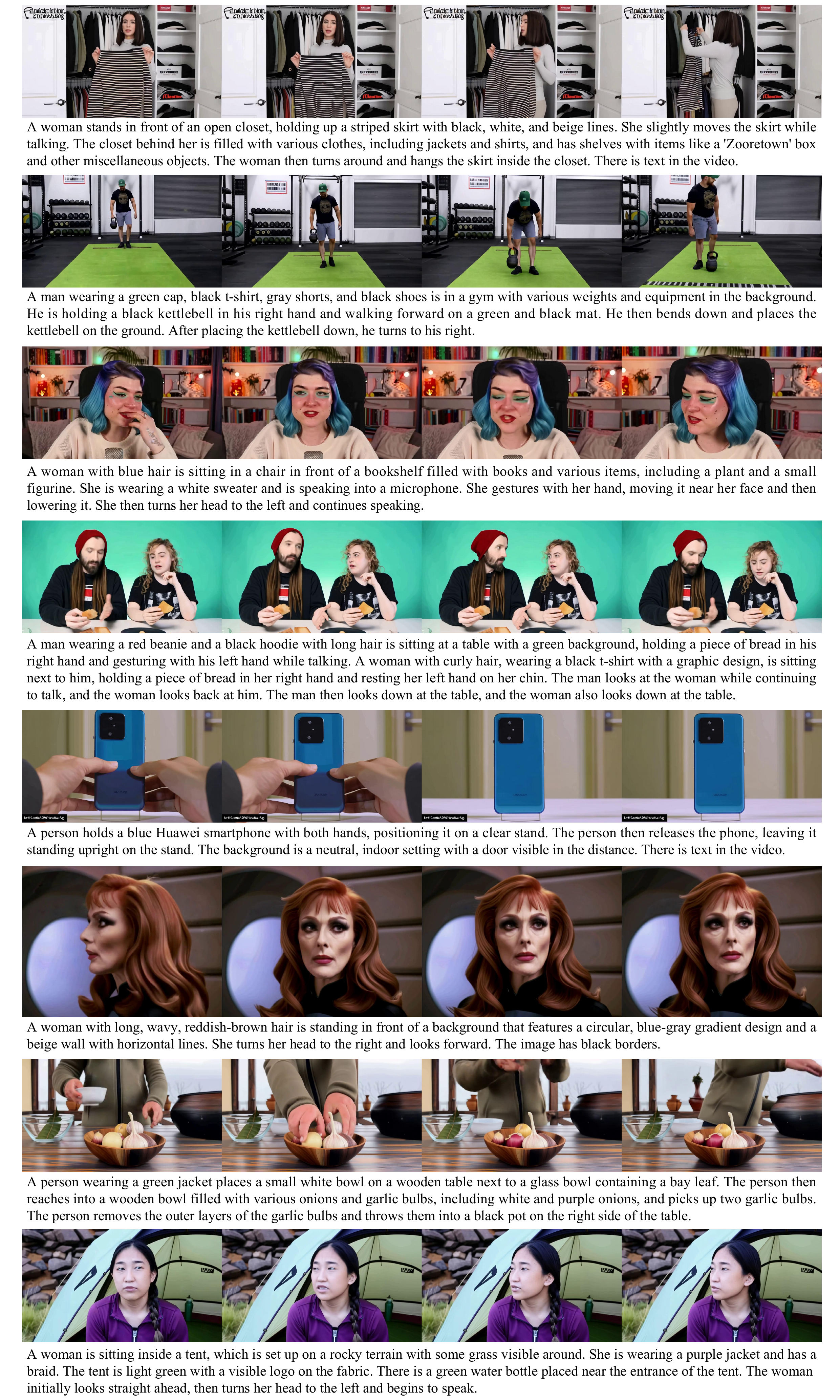}
	\vspace{-0.5cm}
	\caption{\textbf{More qualitative results for the text-to-video generation task.}}
	\vspace{-1.0cm}
	\label{fig:more_t2v_examples_one}
\end{figure}

\begin{figure}[h!]
	\centering
	\vspace{-0.7cm}
	\includegraphics[width=1.0\linewidth]{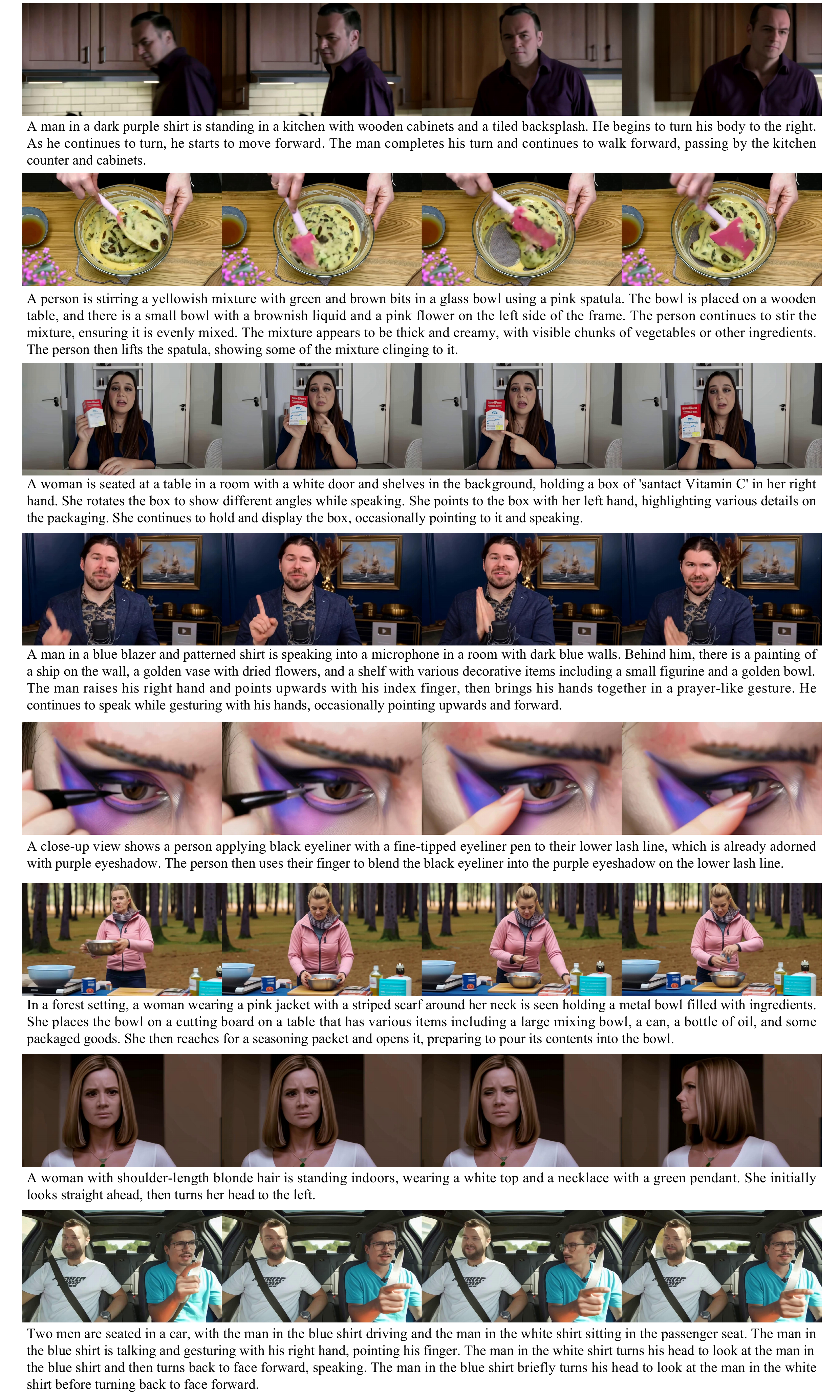}
	\vspace{-0.5cm}
	\caption{\textbf{More qualitative results for the text-to-video generation task.}}
	\vspace{-1.0cm}
	\label{fig:more_t2v_examples_two}
\end{figure}

In Fig. \ref{fig:more_t2v_examples_one} and Fig. \ref{fig:more_t2v_examples_two}, we present more text-to-video generation results of \methodNAME-T2V.

\newpage

